\documentclass[10pt,journal,compsoc]{IEEEtran}

\ifCLASSOPTIONcompsoc
  \usepackage[nocompress]{cite}
\else
  \usepackage{cite}
\fi

\ifCLASSINFOpdf
\else
\fi

\usepackage{times}
\usepackage{epsfig}
\usepackage{graphicx}
\usepackage{amsmath}
\usepackage{amssymb}

\usepackage[utf8]{inputenc} 
\usepackage[T1]{fontenc}    
\usepackage{url}            
\usepackage{booktabs}       
\usepackage{amsfonts}       
\usepackage{nicefrac}       
\usepackage{microtype}      
\usepackage{algorithm}
\usepackage{algorithmic}
\usepackage{array}
\usepackage{bm}
\usepackage{multirow}

\def\obs{{\rm obs}}
\def\syn{{\rm syn}}

\def\I{{\bf I}}
\def\B{{\bf B}}

\def\D{{\cal D}}
\def\S{{\cal S}}

\def\E{{\rm E}}

\def\tI{\tilde{\bf I}}

\def\tM{\tilde{M}}

\begin{document}

\title{Learning Energy-based Spatial-Temporal Generative ConvNets for Dynamic Patterns}

\author{Jianwen Xie,
        Song-Chun Zhu,
        and~Ying Nian Wu
\IEEEcompsocitemizethanks{\IEEEcompsocthanksitem J. Xie is with Hikvision Research Institute, Santa Clara, USA. E-mail: jianwen@ucla.edu. \protect
\IEEEcompsocthanksitem S.-C. Zhu is with the Department of Statistics, University of California, Los Angeles, USA. E-mail: sczhu@stat.ucla.edu. \protect
\IEEEcompsocthanksitem Y. N. Wu is with the Department of Statistics, University of California, Los Angeles, USA. E-mail: ywu@stat.ucla.edu.
}
}

\markboth{}%
{Xie \MakeLowercase{\textit{et al.}}}

\IEEEtitleabstractindextext{%
\begin{abstract}
Video sequences contain rich dynamic patterns, such as dynamic texture patterns that exhibit stationarity in the temporal domain, and action patterns that are non-stationary in either spatial or temporal domain. We show that an energy-based spatial-temporal generative ConvNet can be used to model and synthesize  dynamic patterns. The model defines a probability distribution on the video sequence, and the log probability is defined by a spatial-temporal ConvNet that consists of multiple layers of spatial-temporal filters to capture spatial-temporal patterns of different scales. The model can be learned from the training video sequences by an ``analysis by synthesis'' learning algorithm that iterates the following two steps. Step 1 synthesizes video sequences from the currently learned model. Step 2 then updates the model parameters based on the difference between the synthesized video sequences and the observed training sequences. We show that the learning algorithm can synthesize realistic dynamic patterns. We also show that it is possible to learn the model from incomplete training sequences with either occluded pixels or missing frames, so that model learning and pattern completion can be accomplished simultaneously.
\end{abstract}

\begin{IEEEkeywords}
Deep generative models; Energy-based models; Dynamic textures; Generative ConvNets; Spatial-temporal ConvNets.
\end{IEEEkeywords}}

\maketitle

\IEEEdisplaynontitleabstractindextext

\IEEEpeerreviewmaketitle

\IEEEraisesectionheading{\section{Introduction}\label{sec:introduction}}

\subsection{Background and motivation} 
\IEEEPARstart{T}{here} are a wide variety of dynamic patterns in video sequences, including dynamic textures \cite{doretto2003dynamic} or textured motions \cite{wang2002generative} that  exhibit statistical stationarity or stochastic repetitiveness in the temporal dimension, and action  patterns that are non-stationary in either spatial or temporal domain. 
Recently we have witnessed tremendous
advance in developing discriminative models for dynamic pattern recognition, e.g., \cite{ji20133d}, \cite{karpathy2014large}, \cite{simonyan2014two}, \cite{tran2015learning}, \cite{yue2015beyond}, \cite{wang2016temporal},  \cite{feichtenhofer2017temporal}, and \cite{carreira2017quo}, however, the progress in developing generative models of dynamic patterns for synthesis purpose has been lagging behind. Synthesizing dynamic patterns has been an interesting but challenging problem in computer vision and computer graphics. In this paper, we focus on the task of learning to synthesize dynamic patterns via generative modeling of dynamic patterns with a generative version of the convolutional neural network (ConvNet or CNN), or more specificcally, an energy-based model with ConvNet parametrization of the energy function.

The ConvNet \cite{lecun1998gradient, krizhevsky2012imagenet}  has proven to be an immensely successful discriminative learning machine. The convolution operation in the ConvNet is particularly suited for signals such as images, videos and sounds that exhibit translation invariance either in the spatial domain or the temporal domain or both. Recently, researchers have become increasingly interested in the generative aspects of ConvNet,  for the purpose of visualizing the knowledge learned by the ConvNet, or synthesizing realistic signals, or developing generative models that can be used for unsupervised learning. 

In terms of synthesis,  various approaches based on the ConvNet have been proposed for synthesizing realistic static images \cite{Alexey2015, KarolICML2015, Denton2015a, Kulkarni2015, LuZhuWu2016}. However, there has not been much work in the literature on modeling and synthesizing dynamic patterns based on the ConvNet, and this is the focus of the present paper. 

In the pattern theory \cite{grenander1970unified, grenander2007pattern} of Grenander, a visual pattern is represented by a probability distribution. Grenander used Gibbs distributions or energy-based models to approximate  probability densities for image patterns. In this paper, we continue this paradigm and adopt spatial-temporal convolutional neural networks (ConvNet or CNN) to parametrize the energy functions of the energy-based models that are capable of synthesizing realistic videos of many dynamic patterns.

In terms of generative modeling, generative adversarial networks (GAN) \cite{goodfellow2014generative} and variational autoencoder (VAE) \cite{kingma2013auto} have emerged as two most popular approaches for unsupervised learning of complex distributions. However, neither GAN nor VAE provides explicit probability densities of the data that they model, since they only focus on generating data by learning a mapping from an easily sampled low dimensional distribution (e.g., Gaussian distribution) to the target data distribution. Moreover, both GAN and VAE rely on auxiliary models for training. For example, GAN adopts a discriminator to train the generator in a minimax two-player game, but eventually discards the discriminator after the generator is trained.  VAE recruits an encoder as the inference model to approximate the  inference process based on the posterior distribution, which may cause a gap between the VAE and the maximum likelihood estimator. This paper proposes a different model for dynamic patterns.  It can be learned without recruiting an auxiliary model.

\subsection{Overview of model and algorithm} 
We propose to model dynamic patterns by generalizing the energy-based generative ConvNet model recently proposed by \cite{XieLuICML}. The energy-based generative ConvNet can be derived from the discriminative ConvNet. It is a random field model, a Gibbs distribution, or an energy-based model \cite{lecun2006tutorial, Ng2011} that is in the form of exponential tilting of a reference distribution such as the Gaussian white noise distribution or the uniform distribution. The exponential tilting is parametrized by a ConvNet that involves multiple layers of linear filters and rectified linear units (ReLU)  \cite{krizhevsky2012imagenet}, which seek to capture features or patterns at different scales. 
The log probability density or the energy function of the model is the sum of a ConvNet term that is piecewise linear due to the ReLU non-linearity \cite{montufar2014number} and the $\ell_2$ norm of the signal that comes from the Gaussian white noise reference distribution. As a result, the energy function is piecewise quadratic, and the energy-based generative ConvNet model is piecewise Gaussian. Moreover, the local modes of the distribution are auto-encoding, where the auto-encoding process involves a bottom-up pass that computes the filter responses followed by a top-down pass that reconstructs the signal where the multiple layers of filters in the bottom-up pass serve as the basis functions in the top-down pass. Such an explicit representation is unique among energy-based models  \cite{lecun2006tutorial, Ng2011}, and is a result of the fusion between the ReLU piecewise linear structure and the $\ell_2$ norm term from the Gaussian white noise reference distribution. 

The energy-based generative ConvNet can be sampled by the Langevin dynamics. Because of the aforementioned auto-encoding structure of the energy-based generative ConvNet, the Langevin dynamics is driven by the reconstruction error, i.e., the difference between the current sample and its reconstruction by the above auto-encoding process. 
The model can be learned by the stochastic gradient algorithm  \cite{younes1999convergence}. It is an ``analysis by synthesis'' scheme that seeks to match the synthesized signals generated by the Langevin dynamics to the observed training signals.  Specifically, the learning algorithm iterates the following two steps after initializing the parameters and the synthesized signals. Step 1 updates the synthesized signals by the Langevin dynamics that samples from the currently learned model. Step 2 then updates the parameters based on the difference between the synthesized data and the observed data in order to shift the density of the model from the synthesized data towards the observed data. It is shown by  \cite{XieLuICML}  that the learning algorithm can synthesize realistic spatial image patterns such as textures and objects. 

In this article, we generalize the energy-based spatial generative ConvNet by adding the temporal dimension, so that the resulting ConvNet consists of multiple layers of spatial-temporal filters that seek to capture spatial-temporal patterns at various  scales. 
For dynamic textures, these spatial-temporal filters are convolutional in the temporal domain, reflecting the statistical stationarity or stochastic repetitiveness of the patterns in the temporal domain. If the dynamic textures also exhibit spatial stationarity, we also make the spatial-temporal filters convolutional in the spatial domain. For action and motion patterns that are non-stationary in the temporal domain, the top layer filters are fully connected in the temporal domain. 
 We provide a mode seeking and mode shifting interpretation and an adversarial interpretation of the learning and sampling algorithm. 
We show that the learning algorithm for training the model can synthesize realistic dynamic patterns.  We also show that it is possible to learn the model from incomplete video sequences with either occluded pixels or missing frames, so that model learning and pattern completion can be accomplished simultaneously.

\subsection{Related work} 
In this section, we provide a comprehensive review of related work. 

\textit{Dynamic pattern recognition}. Recognizing dynamic patterns (e.g., actions) with ConvNets has been extensively studied in the past few years. \cite{karpathy2014large} applied ConvNets with deep structures on a large-scale video dataset for classification. \cite{simonyan2014two} designed a two-stream ConvNet structure, which incorporates spatial network trained on still image frames and temporal network trained on motion in the form of dense optical flow, for action recognition. \cite{tran2015learning} trained 3D ConvNets \cite{ji20133d} on the realistic and large-scale video datasets to learn both appearance and motion features with 3D convolution operations. \cite{yue2015beyond} studied modeling long-range temporal structure with ConvNets and LSTM \cite{hochreiter1997long}. \cite{feichtenhofer2016spatiotemporal} and \cite{feichtenhofer2017temporal} generalized the residual networks (ResNets) for the spatiotemporal domain for video action recognition and dynamic scene recognition. \cite{carreira2017quo} proposed the Inception 3D (I3D) model by inflating all the 2D convolution filters and pooling kernels used by the Inception V1 architecture \cite{szegedy2015going} into 3D convolutions and pre-training the model on the large-scale Kinetics human action video dataset \cite{kay2017kinetics}. Instead of learning discriminative spatial-temporal ConvNets of dynamic patterns, our paper mainly focuses on generative modeling of spatial-temporal ConvNets for dynamic pattern synthesis.

\textit{FRAME models}. The FRAME
(Filters, Random fields, And Maximum Entropy) model \cite{zhu1997GRADE, zhu1997minimax, wu2000equivalence} is a Markov random field model (or a Gibbs distribution, or an energy-based model) that defines a probability distribution on the data space. The probability distribution is the maximum entropy distribution that reproduces the statistical properties of filter responses in the observed data. The original FRAME model is a stationary model developed for modeling texture patterns. A non-stationary version of FRAME model designed for object patterns was proposed in \cite{xie2014learning, xie2015boosting}. The filters used in the above two versions of FRAME models are the Gabor filters and the isotropic Difference of Gaussian filters. These are linear filters that capture simple local image features, such as edges and blobs. Inspired by the recent successes of deep convolutional neural networks (CNNs or ConvNets) \cite{lecun1998gradient, krizhevsky2012imagenet}, the deep FRAME model \cite{LuZhuWu2016} replaces the linear filters by the non-linear filters at a certain convolutional layer of a pre-trained deep ConvNet. Such filters can capture more complex patterns, and the deep FRAME model built on such filters can be more expressive. Our model is a spatial-temporal (3D) generalization of the deep FRAME model with the non-linear filters in the deep ConvNet trained by maximum likelihood from the observed data.
 
\textit{Energy-based generative ConvNet}. Recently, a deep generative model directly derived from the discriminative ConvNet model was proposed in \cite{XieLuICML}. The resulting model is an energy-based generative ConvNet model. The maximum likelihood learning of the model involves Markov chain Monte Carlo (MCMC) approximation of the gradient of the data log-likelihood. To address the inefficiency of MCMC sampling, \cite{gao2018learning} developed a multi-grid sampling and learning method for the energy-based generative ConvNet, and \cite{xie2016cooperative} proposed to recruit a top-down generator \cite{HanLu2016} serving as an approximate sampler of the energy-based generative ConvNet model.  \cite{XieLuICML} did not work on dynamic patterns such as those in the video sequences. \cite{xie2017synthesizing} is a generalization of  \cite{XieLuICML} for dynamic patterns by adopting spatial-temporal ConvNets \cite{ji20133d} to capture spatial and temporal features of the video sequences. Recently, \cite{xie2018learning} proposed a volumetric version of the energy-based generative ConvNet for modeling 3D shape patterns. This paper is an expanded version of
our conference paper in \cite{xie2017synthesizing}.

\textit{Dynamic textures}. Generating dynamic textures or textured motions have been studied by \cite{doretto2003dynamic,wang2002generative,wang2004analysis, han2015video}. For instance,  \cite{doretto2003dynamic} proposed a vector auto-regressive model coupled with frame-wise dimension reduction by single value decomposition. It is a linear model with Gaussian innovations. \cite{wang2002generative} proposed a dynamic model based on sparse linear representation of frames. See \cite{you2016kernel} for a recent review of dynamic textures. The spatial-temporal generative ConvNet is a non-linear and non-Gaussian model and is expected to be more flexible in capturing complex spatial-temporal patterns in dynamic textures with multiple layers of non-linear spatial-temporal filters. Recently some researcher have started to study dynamic texture synthesis by matching feature statistics, which are extracted by pre-trained convolutional networks, between synthesized and observed examples, e.g., \cite{funke2017synthesising} adopted a pre-trained ConvNet for object recognition, while \cite{tesfaldet2018two} used two pre-trained ConvNets trained for object recognition and optical flow prediction separately. Even though a ConvNet structure is also included in our model, it serves as the energy function of the energy-based model and is learned from scratch by maximizing the log-likelihood of the observed data,  without relying on other pre-trained networks for assistance.

\textit{Generative adversarial networks of videos}. Recently, multiple video generation frameworks using generative adversarial network (GAN) \cite{goodfellow2014generative} were proposed. For example, one can generalize the existing image-based generative adversarial networks framework to video generation by using a single generator consisting of 3D deconvolutional layers. \cite{vondrick2016generating} proposed  generative adversarial networks for video with a spatial-temporal convolutional architecture that disentangles the scene's foreground from the background. TGAN \cite{saito2017temporal} exploited a 1D temporal generator and a 2D image generator for video generation. The temporal generator takes a single latent variable as input and outputs a set of latent variables, while the image generator transforms these latent variables provided by the temporal generator into video frames. MoCoGAN\cite{tulyakov2018mocogan} proposed the motion and content decomposed generative adversarial networks for video generation. All of the above methods need to recruit a discriminator with appropriate convolutional architecture to evaluate whether the generated videos are from the training data or the video generator. Different from GAN-based methods, our model is a deep 3D convolutional energy-based model with only one single bottom-up 3D ConvNet architecture as the energy function. Our model generates video clips by directly sampling from an explicit distribution by MCMC, such as Langevin dynamics, while the GAN-based methods cannot provide explicit distributions of the videos that they seek to model. Our model  has an adversarial interpretation. See section \ref{sect:adv} for details. 

\textit{Recurrent neural network}. For temporal data, a popular model is the recurrent neural network \cite{williams1989learning, hochreiter1997long}. It is a causal model and it requires a starting frame. In contrast, our model is non-causal, and does not require a starting frame. Compared to the recurrent network, our model is more convenient and direct in capturing temporal patterns at multiple time scales. 

Subsection \ref{sec:c} presents a detailed comparison between various generative models of dynamic patterns that are based on deep neural networks. 

\subsection{Contributions}
The following are the main contributions of this paper. (1) We propose an energy-based spatial-temporal generative ConvNet for modeling video sequences by combining the 3D ConvNets \cite{ji20133d} and the energy-based generative ConvNets \cite{XieLuICML}. (2) We show that the model is capable of synthesizing realistic dynamic patterns. (3) We show that it is possible to learn the model from incomplete data. (4) We present a mode seeking and mode shifting interpretation of the learning and sampling algorithm, and we also present an adversarial interpretation of the algorithm. 

\section{Energy-based Spatial-temporal generative ConvNet}
\label{gen_inst}

\subsection{Spatial-temporal filters}

To fix notation, let $\I(x, t)$ be an image sequence of a video defined on the square (or rectangular) image domain $\D$ and the time domain ${\cal T}$, where $x = (x_1, x_2) \in \D$ indexes the coordinates of pixels, and $t \in {\cal T}$ indexes the frames in the video sequence. We can treat $\I(x, t)$ as a three dimensional function defined on $\D \times {\cal T}$.  For a spatial-temporal filter $F$,  we let $F*\I$ denote the filtered image sequence or feature map, and let $[F*\I](x, t)$ denote the filter response or feature at pixel $x$ and time $t$. 

The spatial-temporal ConvNet is a composition of multiple layers of linear filtering and ReLU non-linearity, as expressed by the following recursive formula: 
\begin{equation}
\begin{aligned}
 & [F^{(l)}_{k}  *\I] (x, t)  =    h\Bigg(\sum_{i=1}^{N_{l-1}}  \sum_{(y, s) \in \S_{l}} w^{(l, k)}_{i, y, s}    \\
 &  \times  [F^{(l-1)}_{i}*\I](x+y, t + s) + b_{l, k}\Bigg), 
\end{aligned}
\label{eq:ConvNet}
\end{equation}
where $l \in \{1, 2, ..., {L}\}$ indexes the layers.  $\{F^{(l)}_k, k = 1, ..., N_l\}$ are the filters at layer $l$, and $\{F^{(l-1)}_i, i = 1, ..., N_{l-1}\}$ are the filters at layer $l-1$.  $k$ and $i$ are used to index filters at layers $l$ and $l-1$ respectively, and $N_l$ and $N_{l-1}$ are the numbers of filters at layers $l$ and $l-1$ respectively. The filters are locally supported, so the range of $(y, s)$  is within a local support $\S_{l}$  (such as a $7 \times 7 \times 3$ box of image sequence). The weight parameters $(w^{(l, k)}_{i, y, s}, (y, s)  \in \S_l, i = 1, ..., N_{l-1})$ define a linear filter that operates on $(F^{(l-1)}_{i}*\I, i = 1, ..., N_{l-1})$. The $\{b_{l,k}\}$ are bias parameters. The linear filtering operation is followed by ReLU $h(r) = \max(0, r)$.  At the bottom layer, $[F^{(0)}_k*\I](x, t) = \I_k(x, t)$, where $k \in \{{\rm R, G, B}\}$ indexes the three color channels. Sub-sampling may be implemented so that in  $[F^{(l)}_{k}  *\I](x, t)$, $x \in \D_l \subset \D$, and $t \in {\cal T}_l \subset {\cal T}$. For example, if the sub-sampling size is $n$, we only keep the first pixel and then every $n$-th pixel after the first. The sub-sampling operation can be applied to both spatial and temporal domains.

The spatial-temporal filters at multiple layers are expected to capture the spatial-temporal patterns at multiple scales. It is possible that the top-layer filters are fully connected in the spatial domain as well as the temporal domain (e.g., the feature maps are $1 \times 1$ in the spatial  domain) if the dynamic pattern does not exhibit spatial or temporal stationarity. 

\subsection{Energy-based spatial-temporal generative ConvNet} 

The spatial-temporal generative ConvNet is an energy-based model or a random field model  defined on the image sequence $\I = (\I(x, t), x \in {\cal D}, t \in {\cal T})$. It is in the form of exponential tilting of a reference distribution $q(\I)$: 
\begin{equation}
p(\I; \theta) = \frac{1}{Z(\theta)} \exp \left[ f(\I; \theta)\right] q(\I), 
\label{eq:ConvNet-FRAME}
\end{equation}
where the scoring function $f(\I; \theta)$ is 
\begin{equation}
f(\I; \theta) =\sum_{k=1}^{K} \sum_{x \in {\cal D}_L} \sum_{t \in {\cal T}_L} [F_k^{(L)}*\I](x, t),  
\label{eq:scoring}
\end{equation}
where $\theta$ consists of all the weight and bias terms that define the filters $(F_k^{(L)}, k = 1, ..., K = N_L)$ at layer $L$, and $q$ is the Gaussian white noise model, i.e., 
\begin{equation}
q(\I) = \frac{1}{(2\pi\sigma^2)^{|{\cal D} \times {\cal T}|/2}} \exp\left[- \frac{1}{2\sigma^2} ||\I||^2\right], 
\label{eq:Gaussian}
\end{equation}   
where $|\D \times {\cal T}|$ counts the number of pixels in the domain $\D \times {\cal T}$. Without loss of generality, we shall assume $\sigma^2 = 1$. $Z(\theta)=\int \exp [ f(\I;\theta)]q(\I)d\I$ is the normalizing constant or partition function that is analytically intractable. 

The scoring function $f(\I; \theta)$ in (\ref{eq:scoring}) tilts the Gaussian reference distribution into a non-Gaussian model. In fact, the purpose of $f(\I; \theta)$ is to identify the non-Gaussian spatial-temporal features or patterns. In the definition of $f(\I; \theta)$ in (\ref{eq:scoring}), we sum over the filter responses at the top layer $L$ over all the filters, positions and times. The spatial and temporal pooling reflects the fact that we assume the model is stationary in spatial and temporal domains. If the dynamic texture is non-stationary in the spatial or temporal domain, then  the top layer filters $F_k^{(L)}$ are fully connected in the spatial or temporal domain, e.g., ${\cal D}_L$ is $1 \times 1$. 

A simple but consequential property of the ReLU non-linearity is that $h(r) = \max(0, r) = 1(r>0) r$, where $1()$ is the indicator function, so that $1(r>0) = 1$ if $r>0$ and $0$ otherwise.  As a result, the scoring function $f(\I; \theta)$ is piecewise linear \cite{montufar2014number}, and each linear piece is defined by the multiple layers of binary activation variables $\delta_{k, x, t}^{(l)}(\I; \theta) = 1\left([F_k^{(l)}*\I](x, t)>0\right)$, which tells us whether a local spatial-temporal pattern represented by the $k$-th filter at layer $l$,  $F_k^{(l)}$, is detected at position $x$ and time $t$. Let $\delta(\I; \theta) = \left(\delta_{k, x, t}^{(l)}(\I; \theta), \forall l, k, x, t \right)$ be the activation pattern of $\I$. Then $\delta(\I; \theta)$ divides the image space into a large number of pieces according to the value of $\delta(\I; \theta)$. On each piece of image space with fixed $\delta(\I; \theta)$, the scoring function $f(\I; \theta)$ is linear, because with fixed value of the indicator function $1(r>0)$, the ReLU $h(r)$ reduces to a linear mapping. Specifically, according to \cite{XieLuICML}, we can write
\begin{equation}
f(\I; \theta) = a_{\theta, \delta(\I; \theta)}  + \langle \I, \B_{\theta, \delta(\I; \theta)}\rangle,
\end{equation}
 where both $a$ and $\B$ are defined by $\delta(\I; \theta)$ and  $\theta$. In fact, $\B = \partial f(\I; \theta)/\partial \I$, and can be computed by back-propagation, with $h'(r) = 1(r>0)$. The back-propagation process defines a top-down deconvolution process \cite{zeiler2011adaptive}, where the filters at multiple layers become the basis functions at those layers, and the activation variables at different layers in $\delta(\I; \theta)$ become the coefficients of the basis functions in the top-down deconvolution. 

$p(\I; \theta)$ in  (\ref{eq:ConvNet-FRAME})  is an energy-based model \cite{lecun2006tutorial, Ng2011}, whose energy function is a combination of the $\ell_2$ norm $\|\I\|^2$ that comes from the reference distribution $q(\I)$ and the piecewise linear scoring function $f(\I; \theta)$, i.e., 
\begin{equation}
\begin{aligned}
 {\cal E}(\I; \theta) &= -f(\I; \theta) + \frac{1}{2} \|\I\|^2\\
 & =\frac{1}{2}  \|\I\|^2 - \left(a_{\theta, \delta(\I; \theta)}  + \langle \I, \B_{\theta, \delta(\I; \theta)}\rangle \right)\\
 &= \frac{1}{2} \|\I - \B_{\theta, \delta(\I; \theta)}\|^2 + {\rm const}, 
\label{eq:ConvNet-FRAME3}
\end{aligned}
\end{equation}
where ${\rm const} = -a_{\theta, \delta(\I; \theta)} - \|\B_{\theta, \delta(\I; \theta)}\|^2/2$, which is constant on the piece of image space with fixed $\delta(\I; \theta)$.

Since ${\cal E}(\I; \theta)$ is a piecewise quadratic function,  $p(\I; \theta)$ is piecewise Gaussian. On the piece of image space $\{\I: \delta(\I; \theta) = \delta\}$, where $\delta$ is a fixed value of $\delta(\I; \theta)$, $p(\I; \theta)$ is ${\rm N}(\B_{\theta, \delta}, {\bf 1})$ truncated to  $\{\I: \delta(\I; \theta) = \delta\}$, where we use ${\bf 1}$ to denote the identity matrix. If the mean of this Gaussian piece,  $\B_{\theta, \delta}$,  is within $\{\I: \delta(\I; \theta) = \delta\}$, then $\B_{\theta, \delta}$ is also a local mode, and this local mode $\I$ satisfies a hierarchical auto-encoder, with a bottom-up encoding process $\delta = \delta(\I; \theta)$, and a top-down decoding process $\I = \B_{\theta, \delta}$. In general, for an image sequence $\I$, $\B_{\theta, \delta(\I; \theta)}$ can be considered a reconstruction of $\I$, and this reconstruction is exact if $\I$ is a local mode of ${\cal E}(\I; \theta)$. 

Our model in (\ref{eq:ConvNet-FRAME}) directly corresponds to a classifier, specifically a spatial-temporal discriminative ConvNet, in the following sense \cite{dai2014generative, XieLuICML, tu2007learning, lazarow2017introspective, jin2017introspective, lee2017wasserstein}. Suppose we have $C$ categories of video sequences. Let $p_c(\I;\theta)$ be the video sequence distribution of category $c$, for $c=1,...,C$, which is defined according to model (\ref{eq:ConvNet-FRAME}). The ConvNet structures $f_c (\I)$ may share common lower layers of parameters. Let $\rho_c$ be the prior probability of category $c$ for $c=1,...,C$. The posterior distribution $p(c|\I)$ for classifying an example $\I$ to the category $c$ is a softmax multi-class classifier (Spatial-temporal discriminative ConvNet) given by
\begin{equation}
p(c|\I;\theta) =  \frac{\exp[f_c(\I;\theta)+b_c]}{\sum_{c=1}^C \exp[f_c(\I; \theta)+b_c]},
\label{eq:discriminative}
\end{equation}
where the category-specific bias term $b_c = \log \rho_c - \log Z_c(\theta) + \text{constant}$. The correspondence to discriminative ConvNet justifies the energy-based spatial-temporal generative ConvNet model.

\subsection{Sampling and learning algorithm}

One can sample from $p(\I; \theta)$  of model (\ref{eq:ConvNet-FRAME}) by the Langevin dynamics: 
\begin{eqnarray}
\I_{\tau+1} &=& \I_{\tau} - \frac{\epsilon^2}{2}  \frac{\partial}{\partial \I}{\cal E}(\I; \theta)    + \epsilon Z_\tau \nonumber \\ 
&=& \I_{\tau} - \frac{\epsilon^2}{2} \left[\I_\tau - \frac{\partial}{\partial \I}f(\I; \theta)\right] + \epsilon Z_\tau \nonumber \\ 
&=& \I_{\tau} - \frac{\epsilon^2}{2} \left[\I_\tau - \B_{\theta, \delta(\I_\tau; \theta)}\right] + \epsilon Z_\tau,
\label{eq:Langevin}
\end{eqnarray}
 where $\tau$ indexes the time steps,  $\epsilon$ is the step size, and  $Z_\tau \sim {\rm N}(0, {\bf 1})$ is the Gaussian white noise term. 
 
The Langevin dynamics consists of a deterministic part and a stochastic part. The former is a gradient descent that attempts to find the minimum of the energy function ${\cal E}(\I; \theta)$, while the latter is a Brownian motion $Z_r$ that prevents the gradient descent from falling into local minimum traps.

 The dynamics is driven by the reconstruction error $\I -\B_{\theta, \delta(\I; \theta)}$. The finiteness of the step size $\epsilon$ can be corrected by a Metropolis-Hastings acceptance-rejection step. The Langevin dynamics can be extended to Hamiltonian Monte Carlo \cite{neal2011mcmc} or more sophisticated versions \cite{girolami2011riemann}.

The learning of $\theta$ from training image sequences $\{\I_m, m = 1, ..., M\}$ can be accomplished by the maximum likelihood estimation (MLE), which follows an ``analysis by synthesis'' scheme. Let $L(\theta) = \sum_{m=1}^{M} \log p(\I_m; \theta)/M$, 
with $p(\I; \theta)$ defined in (\ref{eq:ConvNet-FRAME}), 
\begin{equation}
\frac{\partial L(\theta)}{\partial \theta} = \frac{1}{M} \sum_{m=1}^{M} \frac{\partial}{\partial \theta} f(\I_m; \theta) 
   -  \E_{\theta} \left[ \frac{\partial}{\partial \theta} f(\I; \theta) \right],
\label{eq:gradient_L}
\end{equation}
where the $\E_{\theta}$ denotes the expetation with respect to $p(\I;\theta)$, and the expectation term is due to $\frac{\partial }{\partial \theta} \log Z(\theta)=\E_{\theta}[\frac{\partial}{\partial \theta} f(\I; \theta)]$, which is analytically intractable and has to be approximated by the Monte Carlo samples \cite{younes1999convergence} produced by the Langevin dynamics. Suppose we run $\tilde M$ parallel chains of Langevin dynamics according to (\ref{eq:Langevin}) to synthesize $\tilde{M}$ Monte Carlo samples $\{ \tilde{\I}_m, m=1,...,\tilde{M}\}$, then the Monte Carlo approximation of the gradient of the log-likelihood function in (\ref{eq:gradient_L}) is given by
\begin{equation}
\frac{\partial L(\theta)}{\partial \theta} \approx \frac{1}{M} \sum_{m=1}^{M} \frac{\partial}{\partial \theta} f(\I_m; \theta) 
   -  \frac{1}{\tilde{M}} \sum_{m=1}^{\tilde{M}} \frac{\partial}{\partial \theta} f(\tilde{\I}_m; \theta).
\label{eq:gradient_L2}
\end{equation}
$\theta$ can be updated by a stochastic gradient ascent algorithm \cite{younes1999convergence} with learning rate $\eta$:
\begin{align}
& \theta^{(t+1)} = \theta^{(t)} + \nonumber \\
& \eta_t \left[ \frac{1}{M} \sum_{m=1}^{M} \frac{\partial}{\partial \theta} f(\I_m; \theta^{(t)}) 
   -  \frac{1}{\tilde{M}} \sum_{m=1}^{\tilde{M}} \frac{\partial}{\partial \theta} f(\tilde{\I}_m; \theta^{(t)}) \right]
\label{eq:weight_update}
\end{align}

See Algorithm \ref{code:FRAME}  for a description of the learning and sampling algorithm. The algorithm keeps synthesizing image sequences from the current model, and updating the model parameters in order to match the synthesized image sequences to the observed image sequences. This is an ``analysis by synthesis'' scheme. The convergence of the algorithm has been studied by \cite{robbins1951stochastic, younes1999convergence}.

\begin{algorithm}
\caption{Learning and sampling algorithm}
\label{code:FRAME}
\begin{algorithmic}[1]

\REQUIRE ~~\\
(1)  training image sequences $\{\I_m, m=1,...,M\}$ \\
(2) number of synthesized image sequences $\tilde{M}$\\
(3) number of Langevin steps $l$\\
(4) number of learning iterations $T$

\ENSURE~~\\
(1) estimated model parameters $\theta$\\
(2) synthesized image sequences $\{\tI_m, m = 1, ..., \tilde{M}\}$ 

\item[]
\STATE Let $t\leftarrow 0$, initialize $\theta^{(0)}$.
\STATE Initialize $\tI_m$, for $m = 1, ..., \tilde{M}$, by sampling from $q(\I)$. 
\REPEAT 
\STATE For each $m$, run $l$ steps of Langevin dynamics to update $\tI_m$, i.e., starting from the current $\tI_m$, each step 
follows equation (\ref{eq:Langevin}). 
\STATE Calculate  $H^{\obs} = \sum_{m=1}^{M} \frac{\partial}{\partial \theta} f(\I_m; \theta^{(t)})/M$, and
$H^{\syn} =  \sum_{m=1}^{\tM} \frac{\partial}{\partial \theta} f(\tI_m; \theta^{(t)})/\tM$.
\STATE Update $\theta^{(t+1)} \leftarrow \theta^{(t)} + \eta_t ( H^{\obs} - H^{\syn}) $,  with step size $\eta_t$. 
\STATE Let $t \leftarrow t+1$
\UNTIL $t = T$
\end{algorithmic}
\end{algorithm}

In algorithm \ref{code:FRAME}, the Langevin sampling step involves the computation of $\partial f(\I; \theta)/\partial \I$, and the parameter updating step involves the computation of $\partial f(\I; \theta)/\partial \theta$. Because of the ConvNet structure of $f(\I; \theta)$, both gradients can be computed efficiently by back-propagation, and the two gradients share most of their chain rule computations in back-propagation. The whole learning and sampling algorithm can be considered an alternating back-propagation algorithm: (1) Sampling back-propagation: changing the synthesized examples via Langevin dynamics or gradient descent. (2) Learning back-propagation: changing the model parameters according to the synthesized examples and the observed examples by gradient ascent. In term of MCMC sampling, the Langevin dynamics samples from an evolving distribution because $\theta^{(t)}$ keeps changing. Thus the learning and sampling algorithm runs non-stationary chains. 

\section{Interpretations and related models}
This section presents interpretations of the learning algorithms, as well as related models of dynamic patterns.

\subsection{Adversarial interpretation} \label{sect:adv}

Our spatial-temporal generative ConvNet model is an energy-based model 
\begin{align} 
p(\I; \theta) =\frac{1}{Z(\theta)}  \exp[-{\cal E}(\I; \theta)],
\end{align}
where the energy function ${\cal E}(\I; \theta)= -f(\I; \theta)+\frac{1}{2}\|\I\|^2$.  

 The update of $\theta$ is based on $\partial L(\theta)/\partial \theta$, which can be approximated by 
\begin{eqnarray}
\frac{\partial L(\theta)}{\partial \theta} &\approx& \frac{1}{\tilde{M}} \sum_{m=1}^{\tilde{M}} \frac{\partial}{\partial \theta} {\cal E}(\tilde{\I}_m; \theta)
-\frac{1}{M} \sum_{m=1}^{M} \frac{\partial}{\partial \theta} {\cal E}(\I_m; \theta) \nonumber \\
&=& \frac{\partial}{\partial \theta} \left[\frac{1}{\tilde{M}} \sum_{m=1}^{\tilde{M}}  {\cal E}(\tilde{\I}_m; \theta)
-\frac{1}{M} \sum_{m=1}^{M}  {\cal E}(\I_m; \theta) \right], \nonumber \\ 
\end{eqnarray} 
where $\{\tilde{\I}_m, m = 1, ..., \tilde{M}\}$ are the synthesized image sequences that are generated by the Langevin dynamics. At the zero temperature limit, the Langevin dynamics becomes gradient descent:
\begin{equation}
\tilde{\I}_{\tau+1} = \tilde{\I}_{\tau} - \frac{\epsilon^2}{2} \frac{\partial}{\partial \tilde{\I}} {\cal E}(\tilde{\I}_{\tau}; \theta). \label{eq:Langevin1}
\end{equation}
The algorithm has an adversarial interpretation where the learning and sampling steps play a minimax game. Consider the value function $V(\tilde{\I}_m, m = 1, ..., \tilde{M}; \theta)$: 
 \begin{align}
 \frac{1}{\tilde{M}} \sum_{m=1}^{\tilde{M}} {\cal E}(\tilde{\I}_m; \theta)
 -\frac{1}{M} \sum_{m=1}^{M}  {\cal E}(\I_m; \theta). 
 \end{align}  
The updating of $\theta$ is to increase $V$ by shifting the low energy regions from the synthesized image sequences $\{\tilde{\I}_m\}$ to the observed image sequences $\{\I_m\}$, whereas the updating of $\{\tilde{\I}_m, m = 1, ..., \tilde{M}\}$ is to decrease $V$ by moving the synthesized image sequences towards the low energy regions (i.e., decreasing $ \frac{1}{\tilde{M}} \sum_{m=1}^{\tilde{M}} {\cal E}(\tilde{\I}_m; \theta)$). Therefore, the resulting algorithm approximately solves the minimax problem below 
\begin{align}
\max_\theta \min_{\{ \tilde{\I}_m\}} V(\{ \tilde{\I}_m\}; \theta).
\end{align}  
This is an adversarial interpretation of the learning and sampling algorithm, but the value function is different from that of the generative adversarial learning  \cite{goodfellow2014generative}. 
 It can also be considered a generalization of the herding method \cite{welling2009herding} from exponential family models to general energy-based models. 

In our work, we let $-{\cal E}(\I; \theta) = f(\I; \theta) - \|\I\|^2/2\sigma^2$. We can also let $-{\cal E}(\I; \theta) = f(\I; \theta)$ by assuming a uniform reference distribution $q(\I)$. Our experiments show that the model with the uniform $q$ can also synthesize realistic dynamic patterns.  

\subsection{Mode seeking and mode shifting}

The data distribution $P_{\text{data}}$ might have many local modes, if the observed data are highly varied. Our model parametrized by a ConvNet structure $f(\I;\theta)$ can be flexible and expressive enough to fit such a $P_{\text{data}}$ by creating modes to encode the observed examples with highly diverse patterns. The $f(\I;\theta)$ or equivalently the energy function ${\cal E}(\I; \theta)$ should be learned such that the energy function places lower values on the observed examples than the unobserved examples. This can be achieved by the sampling and learning algorithm presented in Algorithm \ref{code:FRAME}, which can be interpreted as a process that alternates mode seeking and mode shifting. 

The sampling step can be interpreted as mode seeking where the Langevin dynamics searches for low energy (high probability) modes in the landscape defined by ${\cal E}(\I; \theta)$ via stochastic gradient descent and settles the synthesized image sequences $\{ \tilde{\I}_m \}$ around the low energy (high density) regions of ${\cal E}(\I; \theta)$. 

The learning step can be interpreted as mode shifting by shifting the low energy (high density) regions from the synthesized image sequences $\{ \tilde{\I}_m \}$ towards the observed image sequences $\{ \I_m \}$, or shifting the major modes (or basins) of the energy function ${\cal E}(\I; \theta)$ from the synthesized image sequences towards the observed image sequences, until the observed image sequences stay in the major modes of the energy landscape. 

The learning algorithm will create and sharpen the major modes to concentrate on the observed image sequences, if those modes are too diffused around the observed image sequences. The learned energy landscape might have some major modes that are not occupied by the observed examples, and these modes will create examples that are similar to the observed examples.

This mode seeking and mode shifting interpretation is related to Hopfield network \cite{hopfield1982neural} that uses local modes of energy landscape to memorize the observed examples. This is also related to attractor network \cite{seung1998learning} with the Langevin dynamics serving as the attractor dynamics. 

\subsection{Comparison with other models of dynamic patterns} \label{sec:c}

The energy-based model is based on a bottom-up ConvNet $f_\theta(\I)$ defined on the video sequence $\I$. There are other models based on top-down ConvNets. 

One generator model \cite{HanLu2019} assumes a hidden noise vector $z \sim {\rm N}(0, I_d)$, where $I_d$ is the identity matrix of dimension $d$. Then $\I = g(z) + \epsilon$, where $g$ is parametrized by a top-down spatial-temporal ConvNet, and $\epsilon$ is Gaussian white noise.

The other generator model is a non-linear version of the state space model or hidden Markov model, where $s_{t} = r(s_{t-1}, u_t)$, and $\I_t = g(s_t) + \epsilon_t$, where $s_t$ is the state vector, and $u_t$ is the noise vector, and the transition model $r$ is parametrized by a forward network, and the emission model $g$ is parametrized by a top-down network. This model is called dynamic generator model \cite{XieGaoZhengZhuWu2019}. It has a causal structure. It is a non-linear generalization of the dynamic texture model \cite{doretto2003dynamic}. 

The above generator models can be learned by maximum likelihood. They can also be learned jointly with inference models as in VAE, or discriminator models as in GAN, or the energy-based models. In the joint training of the dynamic generator model and the energy-based model, the former plays the role of actor and the latter plays the role of evaluator or critic.

 In comparison with classical mechanics, the energy-based model is similar to the Lagrangian formulation which is based on the action defined on the whole trajectory, whereas the dynamic generator model is similar to the Hamiltonian formulation which unfolds the trajectory over time. 

In the Gibbs distribution in statistical mechanics, the energy function is defined on the spatial domain, and the system evolves over time and converges to the Gibbs distribution \cite{landau2014guide}. In the energy-based model studied in this paper, the energy function is defined on the spatial-temporal domain, where the time dimension is treated in the same way as the spatial dimensions. The Langevin dynamics evolves this spatial-temporal model over a virtue time (denoted by $\tau$).  This treatment is similar to the lattice field theory in physics, which is  also based on a Gibbs distribution whose energy is defined on the spatial-temporal domain, and the MCMC sampler evolves the system over a virtue time \cite{landau2014guide}. 

In the context of image-based control or intuitive physics, we can jointly model the actions and the video sequences, using the energy-based model and the dynamic generator model. The energy function in the energy-based model can be interpreted as the cost function for optimal control, and the dynamic generator model consists of the policy and the dynamics \cite{ziebart2008maximum}. 

Beside the energy-based models and the generator models, there are also flow-based models \cite{kumar2019videoflow}, where the log-likelihoods can be computed in closed form. Specifically, $\I = g(z)$, where $z$ is a noise vector that is of the same dimensionality as $\I$, and $g$ is a composition of a sequence of invertible transformations that are designed such that the inverse and the Jacobian of $g$ can be computed in closed form. 

\section{Experiments}

We learn the energy-based spatial-temporal generative ConvNet from video clips collected from DynTex++ dataset of \cite{ghanem2010maximum} and the Internet. 
We show the synthesis results by displaying the frames in the video sequences. We have posted the synthesis results on the project page \url{http://www.stat.ucla.edu/~jxie/STGConvNet/STGConvNet.html}, so that the reader can watch the videos.

\subsection{Experiment 1: Generating dynamic textures with both spatial and temporal stationarity}

\begin{figure}[h]
\begin{center}
\includegraphics[width=.13\linewidth]{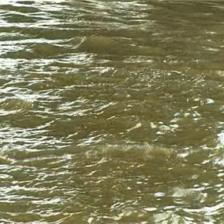}
\includegraphics[width=.13\linewidth]{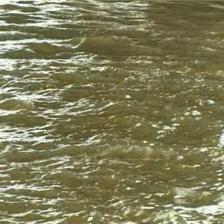}
\includegraphics[width=.13\linewidth]{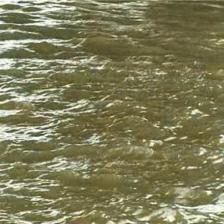}
\includegraphics[width=.13\linewidth]{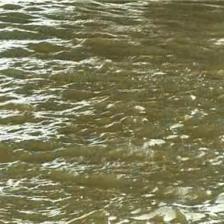}
\includegraphics[width=.13\linewidth]{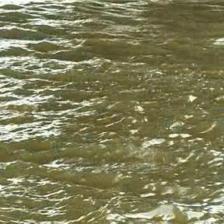}
\includegraphics[width=.13\linewidth]{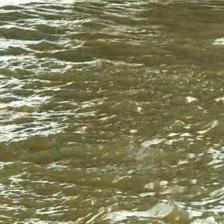}
\includegraphics[width=.13\linewidth]{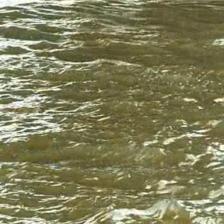}\\[3px]

\includegraphics[width=.13\linewidth]{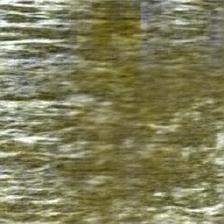}
\includegraphics[width=.13\linewidth]{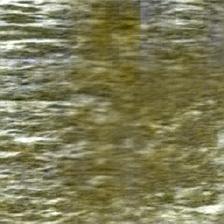}
\includegraphics[width=.13\linewidth]{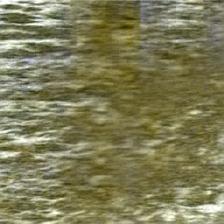}
\includegraphics[width=.13\linewidth]{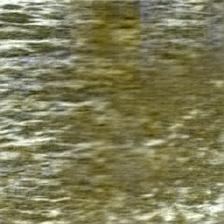}
\includegraphics[width=.13\linewidth]{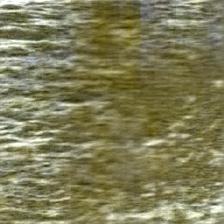}
\includegraphics[width=.13\linewidth]{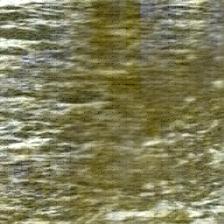}
\includegraphics[width=.13\linewidth]{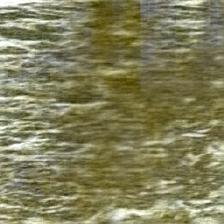} \\[3px]	

\includegraphics[width=.13\linewidth]{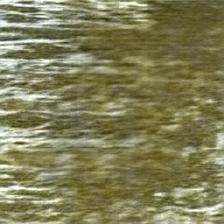}
\includegraphics[width=.13\linewidth]{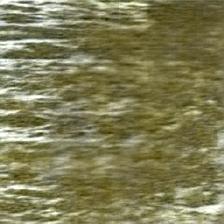}
\includegraphics[width=.13\linewidth]{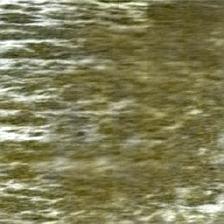}
\includegraphics[width=.13\linewidth]{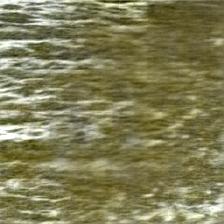}
\includegraphics[width=.13\linewidth]{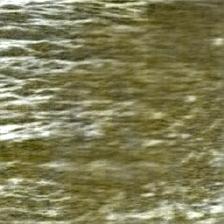}
\includegraphics[width=.13\linewidth]{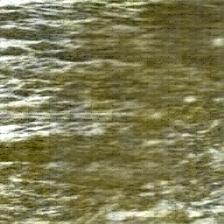}
\includegraphics[width=.13\linewidth]{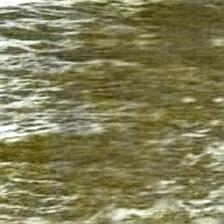}  \\ (a) water wave 1	\\[8px]

\includegraphics[width=.13\linewidth]{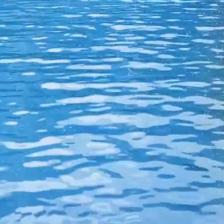}
\includegraphics[width=.13\linewidth]{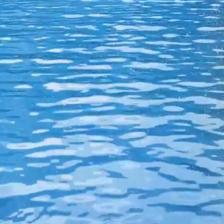}
\includegraphics[width=.13\linewidth]{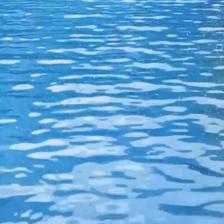}
\includegraphics[width=.13\linewidth]{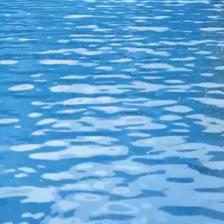}
\includegraphics[width=.13\linewidth]{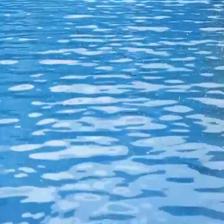}
\includegraphics[width=.13\linewidth]{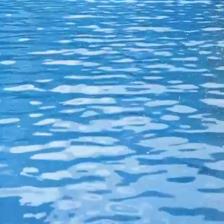}
\includegraphics[width=.13\linewidth]{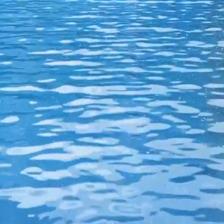}\\
[3px]

\includegraphics[width=.13\linewidth]{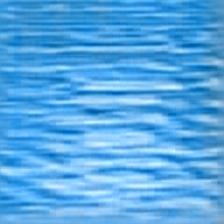}
\includegraphics[width=.13\linewidth]{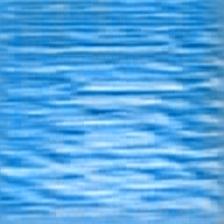}
\includegraphics[width=.13\linewidth]{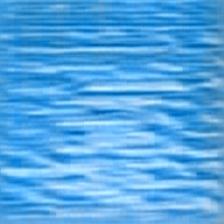}
\includegraphics[width=.13\linewidth]{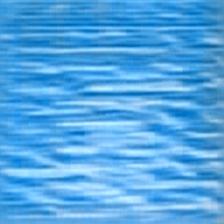}
\includegraphics[width=.13\linewidth]{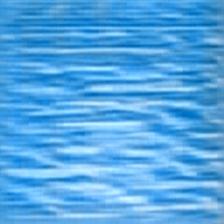}
\includegraphics[width=.13\linewidth]{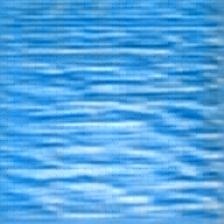}
\includegraphics[width=.13\linewidth]{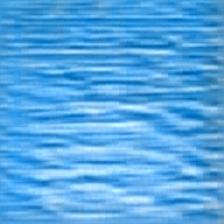}\\[3px]

\includegraphics[width=.13\linewidth]{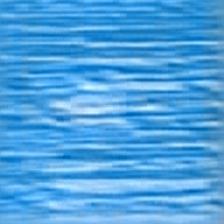}
\includegraphics[width=.13\linewidth]{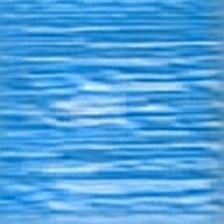}
\includegraphics[width=.13\linewidth]{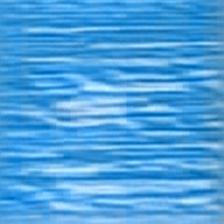}
\includegraphics[width=.13\linewidth]{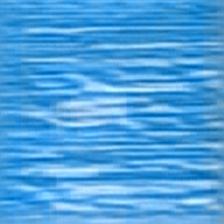}
\includegraphics[width=.13\linewidth]{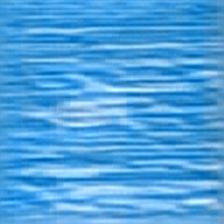}
\includegraphics[width=.13\linewidth]{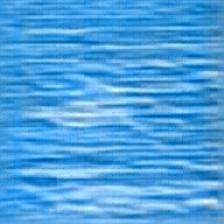}
\includegraphics[width=.13\linewidth]{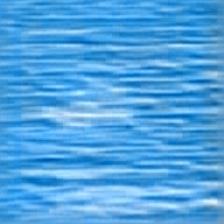}  \\ (b) water wave 2	\\[8px]	
\includegraphics[width=.13\linewidth]{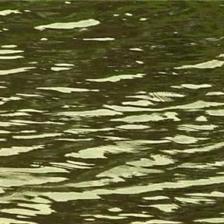}
\includegraphics[width=.13\linewidth]{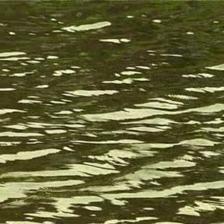}
\includegraphics[width=.13\linewidth]{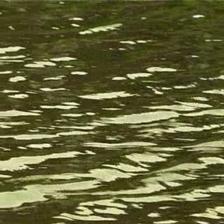}
\includegraphics[width=.13\linewidth]{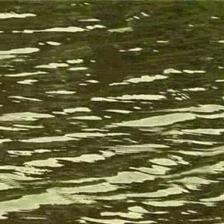}
\includegraphics[width=.13\linewidth]{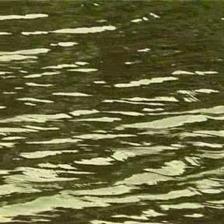}
\includegraphics[width=.13\linewidth]{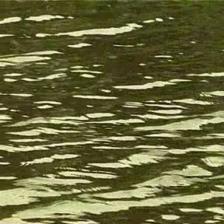}
\includegraphics[width=.13\linewidth]{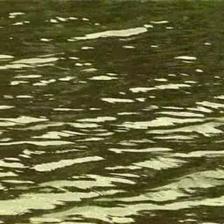} \\[3px]

\includegraphics[width=.13\linewidth]{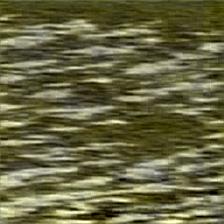}
\includegraphics[width=.13\linewidth]{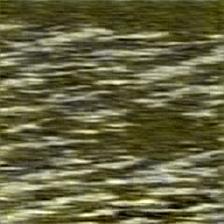}
\includegraphics[width=.13\linewidth]{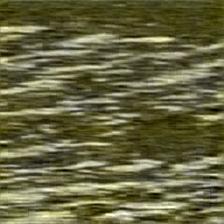}
\includegraphics[width=.13\linewidth]{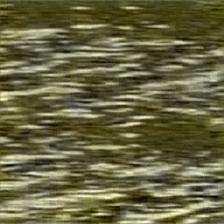}
\includegraphics[width=.13\linewidth]{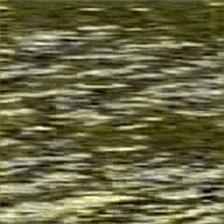}
\includegraphics[width=.13\linewidth]{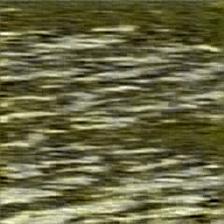}
\includegraphics[width=.13\linewidth]{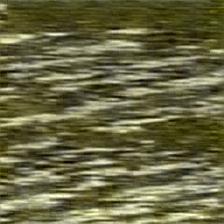} \\[3px]

\includegraphics[width=.13\linewidth]{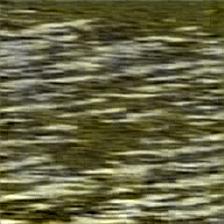}
\includegraphics[width=.13\linewidth]{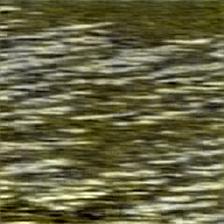}
\includegraphics[width=.13\linewidth]{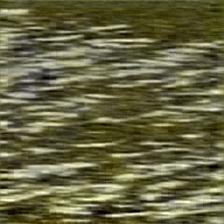}
\includegraphics[width=.13\linewidth]{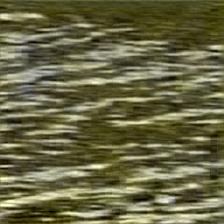}
\includegraphics[width=.13\linewidth]{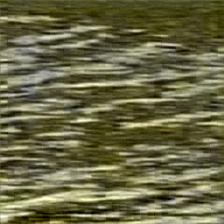}
\includegraphics[width=.13\linewidth]{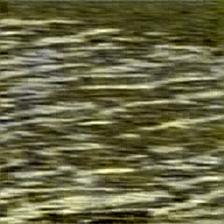}
\includegraphics[width=.13\linewidth]{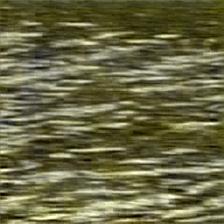}
\\ (c) water wave 3 \\[8px]	

\includegraphics[width=.13\linewidth]{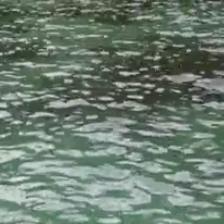}
\includegraphics[width=.13\linewidth]{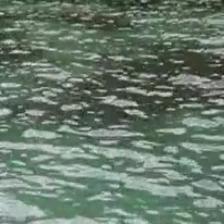}
\includegraphics[width=.13\linewidth]{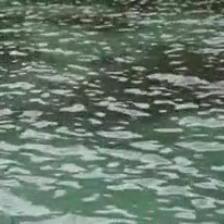}
\includegraphics[width=.13\linewidth]{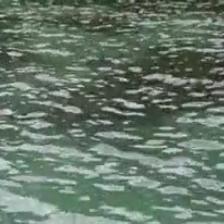}
\includegraphics[width=.13\linewidth]{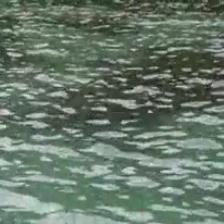}
\includegraphics[width=.13\linewidth]{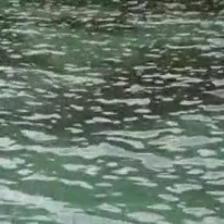}
\includegraphics[width=.13\linewidth]{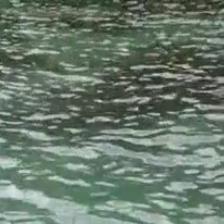} \\[3px]

\includegraphics[width=.13\linewidth]{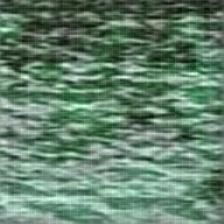}
\includegraphics[width=.13\linewidth]{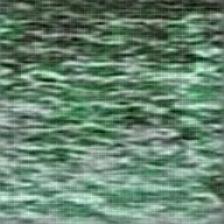}
\includegraphics[width=.13\linewidth]{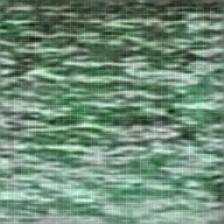}
\includegraphics[width=.13\linewidth]{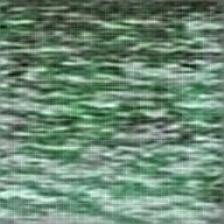}
\includegraphics[width=.13\linewidth]{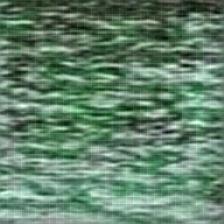}
\includegraphics[width=.13\linewidth]{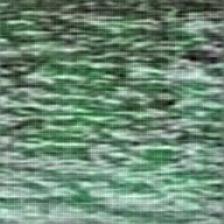}
\includegraphics[width=.13\linewidth]{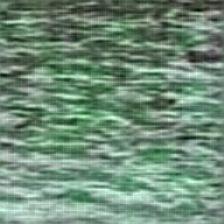} \\[3px]

\includegraphics[width=.13\linewidth]{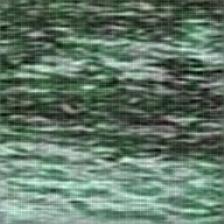}
\includegraphics[width=.13\linewidth]{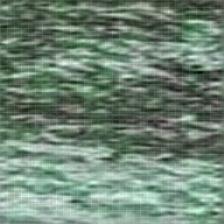}
\includegraphics[width=.13\linewidth]{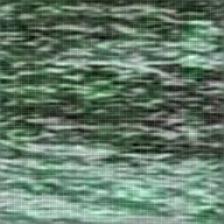}
\includegraphics[width=.13\linewidth]{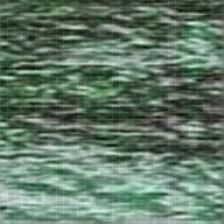}
\includegraphics[width=.13\linewidth]{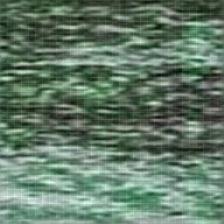}
\includegraphics[width=.13\linewidth]{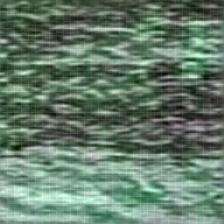}
\includegraphics[width=.13\linewidth]{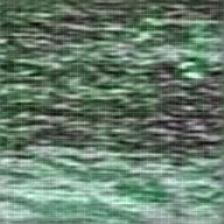}
\\ (d) water wave 4\\
	\caption{Synthesizing dynamic textures with both spatial and temporal stationarity. For each category, the first row displays the frames of the observed sequence, and the second and third rows display the corresponding frames of two synthesized sequences generated by the learning algorithm. The observed and the synthesized videos are of the size 224 $\times$ 224 pixels $\times$ 50 or 70 frames.}
	\label{fig:DTresults1}
\end{center}
\end{figure}

We first learn the model from dynamic textures that are stationary in both spatial and temporal domains. Because these dynamic textures exhibit spatial-temporal homogeneity, we use spatial-temporal filters that are convolutional in both spatial and temporal domains to design our network $f(\I;\theta)$, which consists of three layers of spatial-temporal convolution, ReLU nonlinearity, and sub-sampling. The first layer has 120 $15 \times 15 \times 15$ filters with sub-sampling size of 7 pixels and frames. The second layer has 40 $7 \times 7 \times 7$ filters with sub-sampling size of 3.  The third layer has 20 $3 \times 3 \times 2$ filters with sub-sampling size of $2 \times 2 \times 1$. The sub-sampling operation is useful to reduce the size of the output feature map of each layer for the sake of efficient computation. 

Figure \ref{fig:DTresults1} displays 4 results of learning to synthesize water waves. For each category, the first row displays 7 frames of the observed sequence, while the second and third rows show the corresponding frames of two synthesized sequences generated by the learning algorithm. The qualitative results displayed in Figure \ref{fig:DTresults1} clearly show that our model can learn to generate realistic examples of dynamic textures with both spatial and temporal stationarity, such as water waves. The synthesized examples are similar, but not identical, to the observed video sequence. 
Generally, when the number of chains of synthesis increases, the diversity of the synthesized sequences will also increase.    

We use the layer-by-layer learning scheme. Starting from the first layer, we sequentially add the layers one by one. Each time we learn the model and generate the synthesized image sequence using Algorithm \ref{code:FRAME}. While learning the new layer of filters, we refine the lower layers of filters with back-propagation. Due to the limitation of the GPU memory and for the sake of computational efficiency, we use $\tilde{M}=3$ chains for Langevin sampling. Even with a single chain, the statistics can still be estimated because of stationarity in the temporal and spatial domains. 
We learn an energy-based spatial-temporal generative ConvNet for each category from one observed video that is prepared to be of the size 224 $\times$ 224 pixels $\times$ 50 or 70 frames. The range of pixel intensities is  [0, 255]. Mean subtraction is used as pre-processing. The number of Langevin iterations between every two consecutive updates of parameters, $l = 20$. The number of learning iterations $T = 1200$, where we add one more layer every $400$ iterations.  We use layer-specific learning rates, where the learning rate at the higher layer is less than that at the lower layer, in order to obtain stable convergence.

\subsection{Experiment 2: Generating dynamic textures with only temporal stationarity}

Many dynamic textures have structured background and objects that are not stationary in the spatial domain. In this case, the network used in Experiment 1 may fail. However, we can modify the network in Experiment 1  by using  filters that are fully connected in the spatial domain at the second layer to account for spatial non-homogeneity of the dynamic textures. Specifically, the first layer has 120 $7 \times 7 \times 7$ filters with sub-sampling size of 3 pixels and frames. The second layer is a spatially fully connected layer, which contains 30 filters that are fully connected in the spatial domain but convolutional in the temporal domain. The temporal size of the filters is 4 frames with sub-sampling size of 2 frames in the temporal dimension.  Due to the spatial full connectivity at the second layer, the spatial domain of the feature maps at the third layer is reduced to $1 \times 1$.  The third layer has 5 $1 \times 1 \times 2$ filters with sub-sampling size of 1 in the temporal dimension. 

We  use  end-to-end learning scheme to learn the above 3-layer energy-based spatial-temporal generative ConvNet for dynamic textures with only temporal stationarity. At each iteration, the 3 layers of filters are updated with 3 different layer-specific learning rates. The learning rate at the higher layer is much less than that at the lower layer to avoid the issue of large gradients. 

\begin{figure*}
\begin{center}
\includegraphics[width=.076\linewidth]{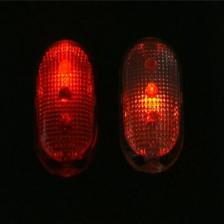}
\includegraphics[width=.076\linewidth]{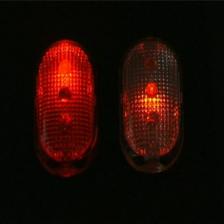}
\includegraphics[width=.076\linewidth]{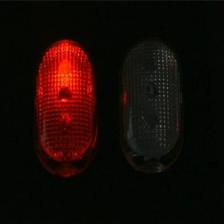}
\includegraphics[width=.076\linewidth]{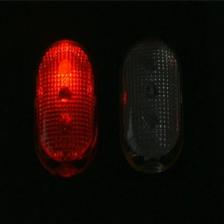} 
\includegraphics[width=.076\linewidth]{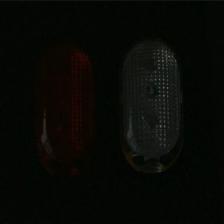} 
\includegraphics[width=.076\linewidth]{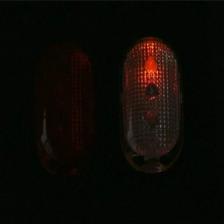}
\hspace{3mm}
\includegraphics[width=.076\linewidth]{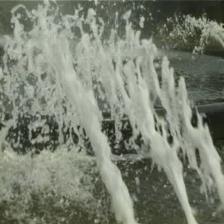}
\includegraphics[width=.076\linewidth]{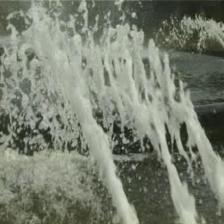}
\includegraphics[width=.076\linewidth]{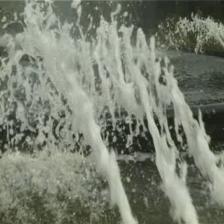}
\includegraphics[width=.076\linewidth]{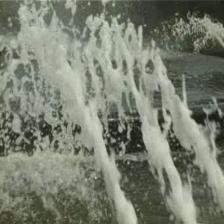}
\includegraphics[width=.076\linewidth]{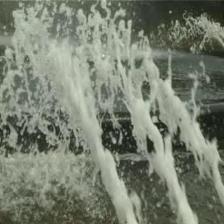}
\includegraphics[width=.076\linewidth]{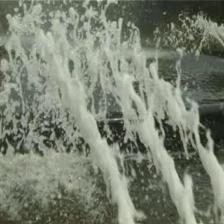} \\[3px]

\includegraphics[width=.076\linewidth]{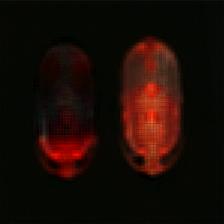}
\includegraphics[width=.076\linewidth]{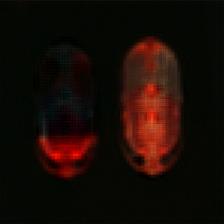}
\includegraphics[width=.076\linewidth]{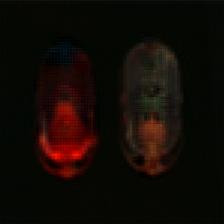}
\includegraphics[width=.076\linewidth]{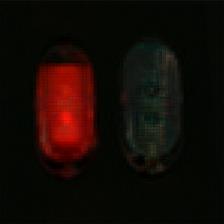} 
\includegraphics[width=.076\linewidth]{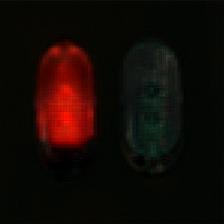} 
\includegraphics[width=.076\linewidth]{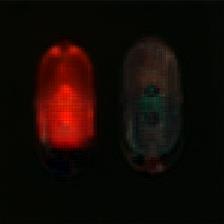}
\hspace{3mm}
\includegraphics[width=.076\linewidth]{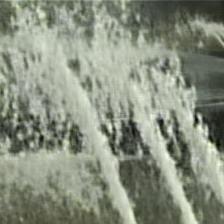}
\includegraphics[width=.076\linewidth]{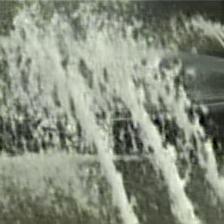}
\includegraphics[width=.076\linewidth]{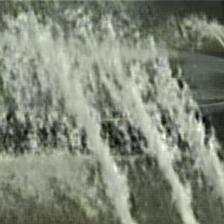}
\includegraphics[width=.076\linewidth]{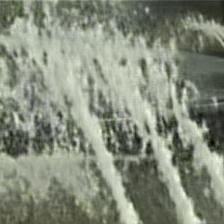}
\includegraphics[width=.076\linewidth]{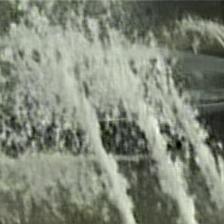}
\includegraphics[width=.076\linewidth]{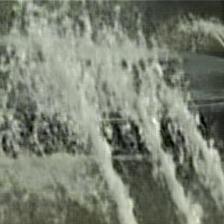}
\\[3px]

\includegraphics[width=.076\linewidth]{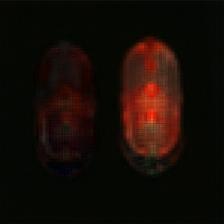}
\includegraphics[width=.076\linewidth]{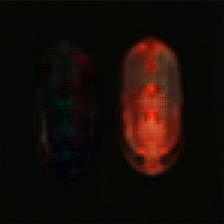}
\includegraphics[width=.076\linewidth]{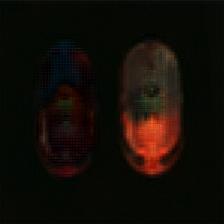}
\includegraphics[width=.076\linewidth]{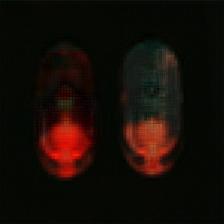} 
\includegraphics[width=.076\linewidth]{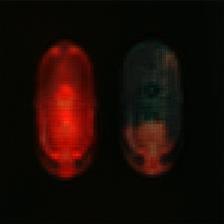} 
\includegraphics[width=.076\linewidth]{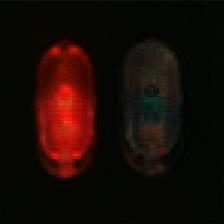}	
\hspace{3mm}
\includegraphics[width=.076\linewidth]{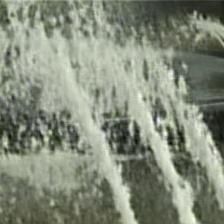}
\includegraphics[width=.076\linewidth]{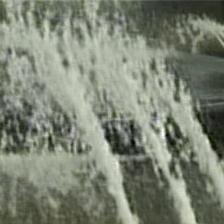}
\includegraphics[width=.076\linewidth]{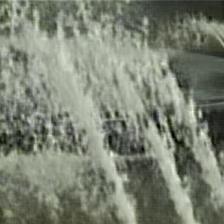}
\includegraphics[width=.076\linewidth]{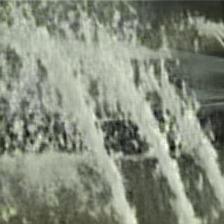}
\includegraphics[width=.076\linewidth]{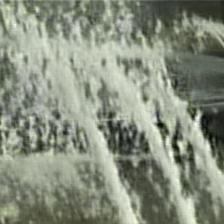}
\includegraphics[width=.076\linewidth]{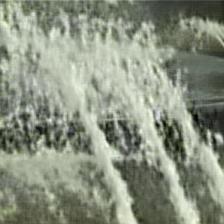}
\\ (a) flashing lights \hspace{75mm} (b) water spray
\\[8px]
\includegraphics[width=.076\linewidth]{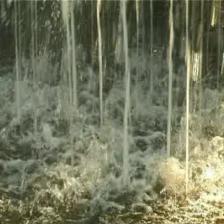}
\includegraphics[width=.076\linewidth]{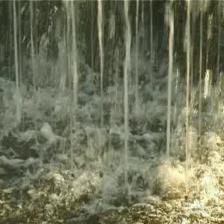}
\includegraphics[width=.076\linewidth]{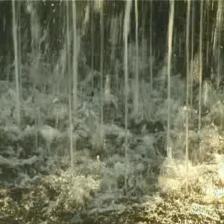}
\includegraphics[width=.076\linewidth]{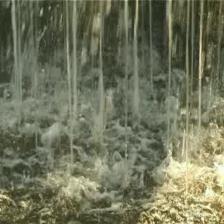}
\includegraphics[width=.076\linewidth]{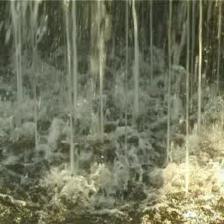} 
\includegraphics[width=.076\linewidth]{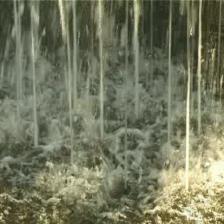} 
\hspace{3mm}
\includegraphics[width=.076\linewidth]{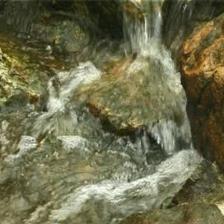}
\includegraphics[width=.076\linewidth]{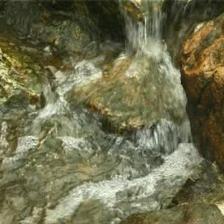}
\includegraphics[width=.076\linewidth]{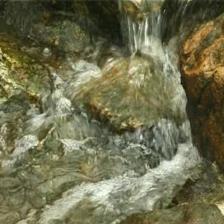}
\includegraphics[width=.076\linewidth]{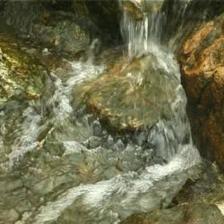}
\includegraphics[width=.076\linewidth]{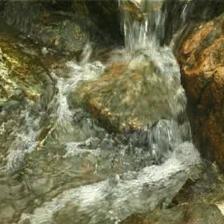}
\includegraphics[width=.076\linewidth]{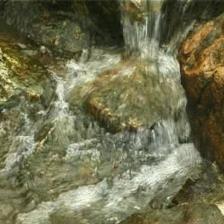}
\\ [3px]
\includegraphics[width=.076\linewidth]{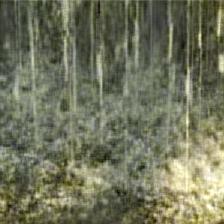}
\includegraphics[width=.076\linewidth]{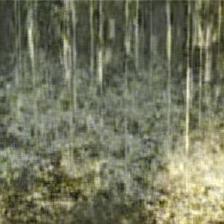}
\includegraphics[width=.076\linewidth]{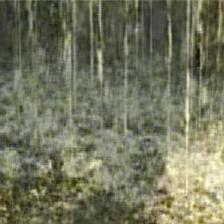}
\includegraphics[width=.076\linewidth]{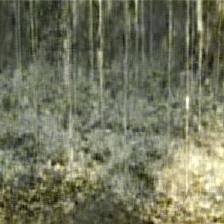}
\includegraphics[width=.076\linewidth]{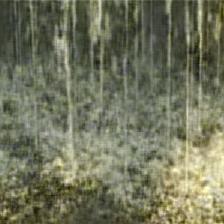} 
\includegraphics[width=.076\linewidth]{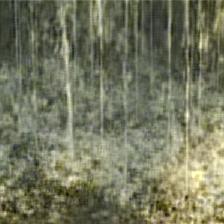} 
\hspace{3mm}
\includegraphics[width=.076\linewidth]{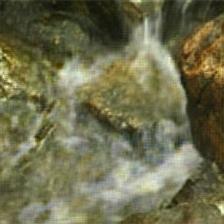}
\includegraphics[width=.076\linewidth]{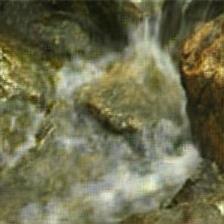}
\includegraphics[width=.076\linewidth]{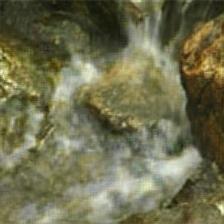}	
\includegraphics[width=.076\linewidth]{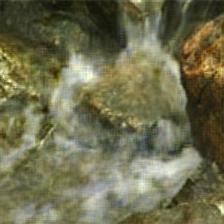}
\includegraphics[width=.076\linewidth]{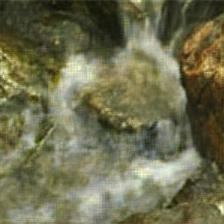}
\includegraphics[width=.076\linewidth]{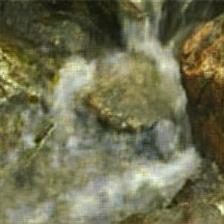} \\ [3px]

\includegraphics[width=.076\linewidth]{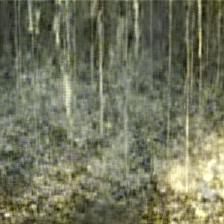}
\includegraphics[width=.076\linewidth]{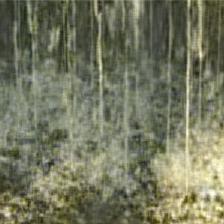}
\includegraphics[width=.076\linewidth]{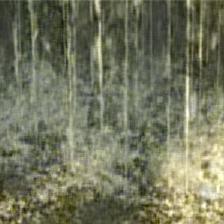}
\includegraphics[width=.076\linewidth]{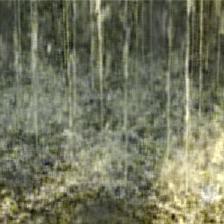}
\includegraphics[width=.076\linewidth]{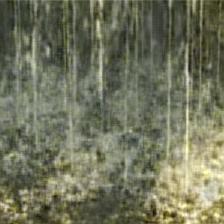} 
\includegraphics[width=.076\linewidth]{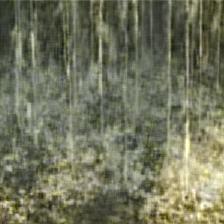}
\hspace{3mm}
\includegraphics[width=.076\linewidth]{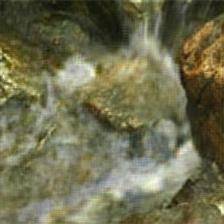}
\includegraphics[width=.076\linewidth]{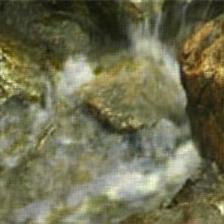}
\includegraphics[width=.076\linewidth]{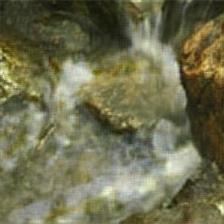}	
\includegraphics[width=.076\linewidth]{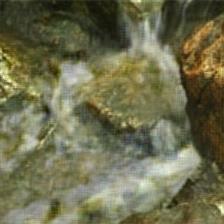}
\includegraphics[width=.076\linewidth]{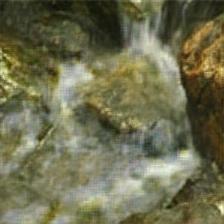}
\includegraphics[width=.076\linewidth]{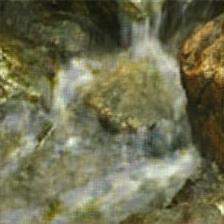}
\\ (c) fountain \hspace{75mm} (d) spring water\\[8px]
\includegraphics[width=.076\linewidth]{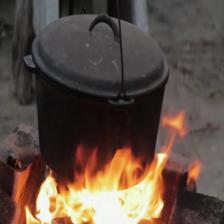}
\includegraphics[width=.076\linewidth]{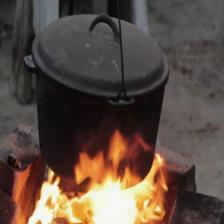}
\includegraphics[width=.076\linewidth]{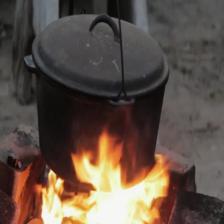}
\includegraphics[width=.076\linewidth]{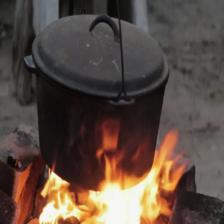}
\includegraphics[width=.076\linewidth]{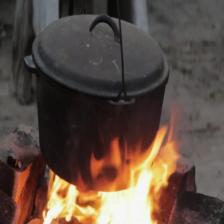} 
\includegraphics[width=.076\linewidth]{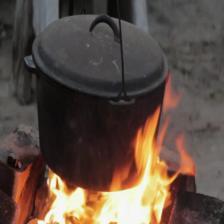}
\hspace{3mm}
\includegraphics[width=.076\linewidth]{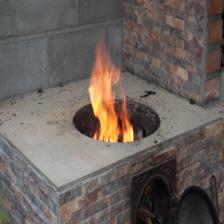}
\includegraphics[width=.076\linewidth]{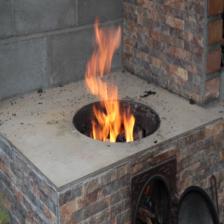}
\includegraphics[width=.076\linewidth]{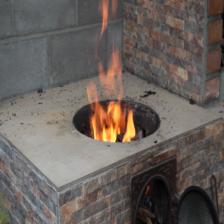}
\includegraphics[width=.076\linewidth]{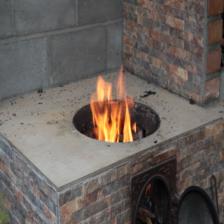}
\includegraphics[width=.076\linewidth]{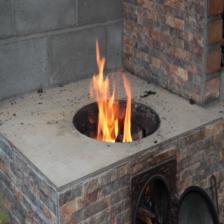}
\includegraphics[width=.076\linewidth]{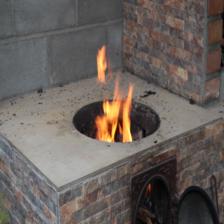}
\\[3px]

\includegraphics[width=.076\linewidth]{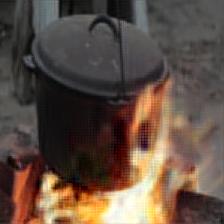}
\includegraphics[width=.076\linewidth]{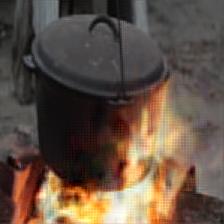}
\includegraphics[width=.076\linewidth]{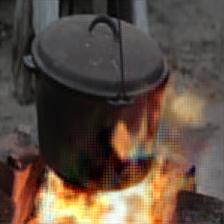}
\includegraphics[width=.076\linewidth]{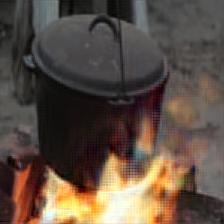}	
\includegraphics[width=.076\linewidth]{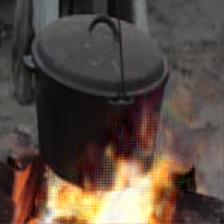}
\includegraphics[width=.076\linewidth]{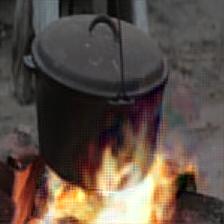}
\hspace{3mm}
\includegraphics[width=.076\linewidth]{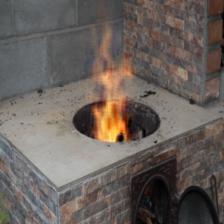}
\includegraphics[width=.076\linewidth]{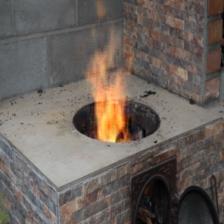}
\includegraphics[width=.076\linewidth]{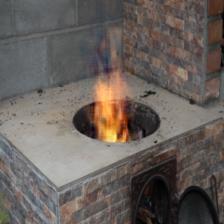}
\includegraphics[width=.076\linewidth]{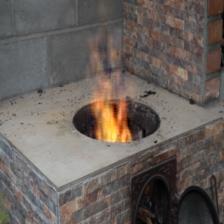}
\includegraphics[width=.076\linewidth]{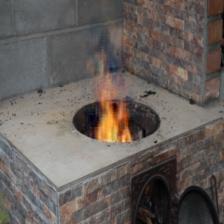}
\includegraphics[width=.076\linewidth]{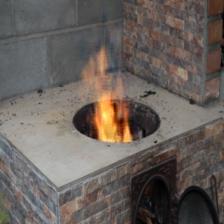}
\\[3px]
\includegraphics[width=.076\linewidth]{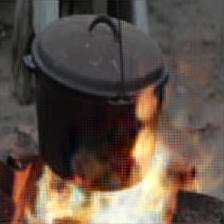}
\includegraphics[width=.076\linewidth]{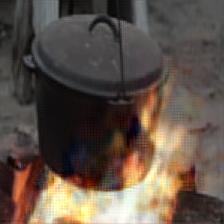}
\includegraphics[width=.076\linewidth]{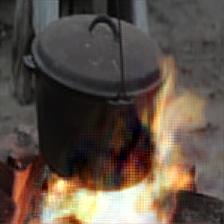}
\includegraphics[width=.076\linewidth]{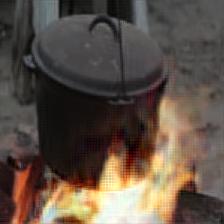}	
\includegraphics[width=.076\linewidth]{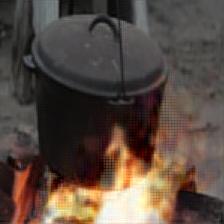}
\includegraphics[width=.076\linewidth]{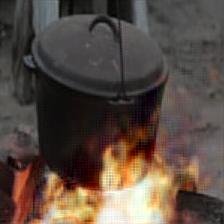}
\hspace{3mm}
\includegraphics[width=.076\linewidth]{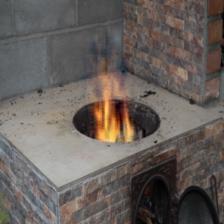}
\includegraphics[width=.076\linewidth]{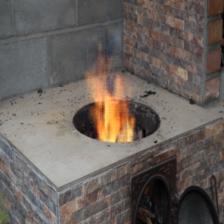}
\includegraphics[width=.076\linewidth]{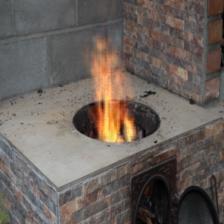}
\includegraphics[width=.076\linewidth]{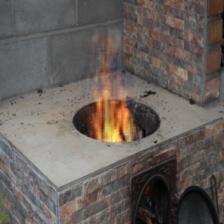}
\includegraphics[width=.076\linewidth]{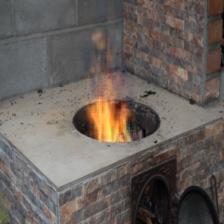}
\includegraphics[width=.076\linewidth]{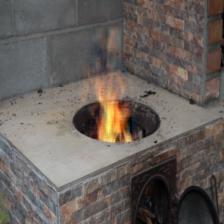}\\
(e) burning fire heating a pot \hspace{58mm} (f) burning fire in a stove\\[8px]
\includegraphics[width=.076\linewidth]{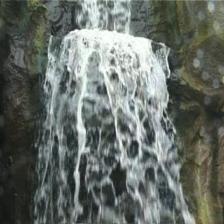}
\includegraphics[width=.076\linewidth]{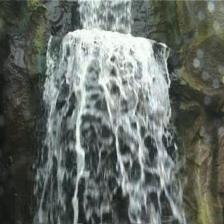}
\includegraphics[width=.076\linewidth]{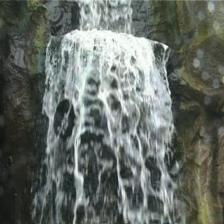}
\includegraphics[width=.076\linewidth]{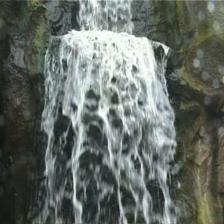}
\includegraphics[width=.076\linewidth]{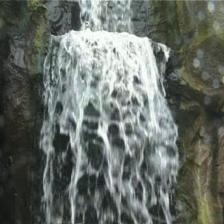} 
\includegraphics[width=.076\linewidth]{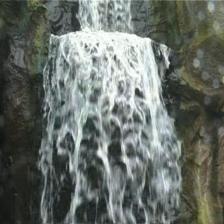}
\hspace{3mm}
\includegraphics[width=.076\linewidth]{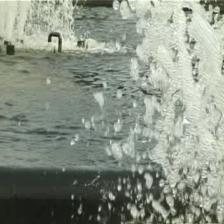}
\includegraphics[width=.076\linewidth]{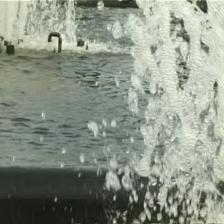}
\includegraphics[width=.076\linewidth]{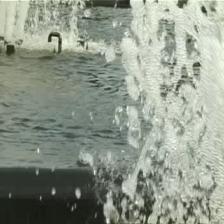}
\includegraphics[width=.076\linewidth]{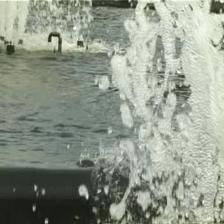}
\includegraphics[width=.076\linewidth]{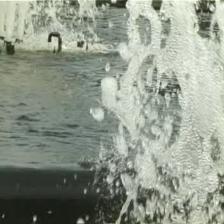}
\includegraphics[width=.076\linewidth]{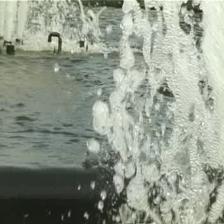}
\\ [3px]
\includegraphics[width=.076\linewidth]{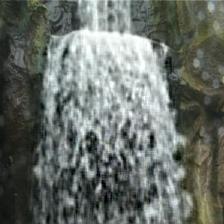}
\includegraphics[width=.076\linewidth]{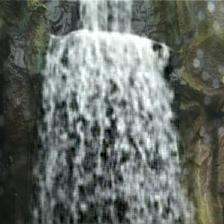}
\includegraphics[width=.076\linewidth]{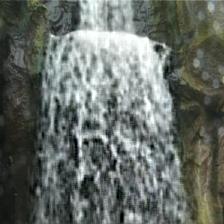}
\includegraphics[width=.076\linewidth]{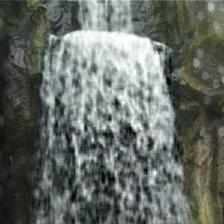}
\includegraphics[width=.076\linewidth]{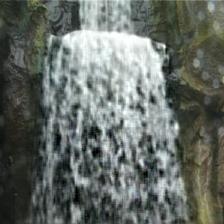}
\includegraphics[width=.076\linewidth]{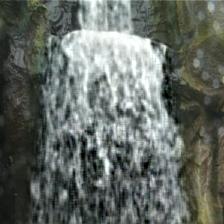}
\hspace{3mm}
\includegraphics[width=.076\linewidth]{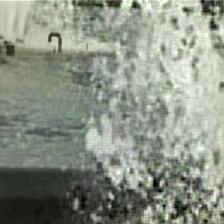}
\includegraphics[width=.076\linewidth]{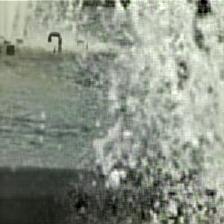}
\includegraphics[width=.076\linewidth]{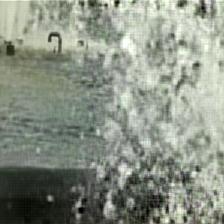}
\includegraphics[width=.076\linewidth]{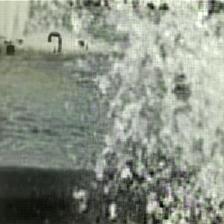}
\includegraphics[width=.076\linewidth]{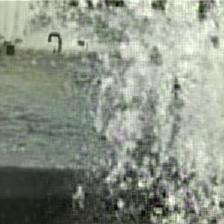}
\includegraphics[width=.076\linewidth]{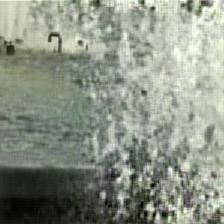}
\\ [3px]
\includegraphics[width=.076\linewidth]{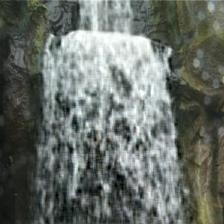}
\includegraphics[width=.076\linewidth]{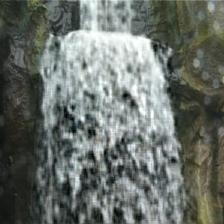}
\includegraphics[width=.076\linewidth]{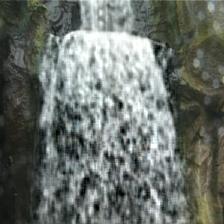}
\includegraphics[width=.076\linewidth]{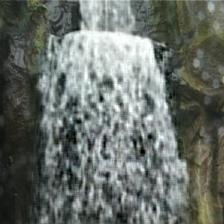}
\includegraphics[width=.076\linewidth]{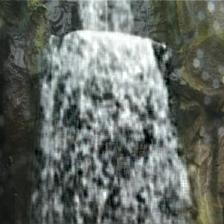}
\includegraphics[width=.076\linewidth]{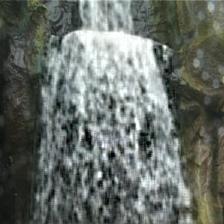}
\hspace{3mm}
\includegraphics[width=.076\linewidth]{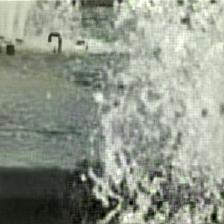}
\includegraphics[width=.076\linewidth]{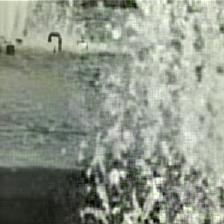}
\includegraphics[width=.076\linewidth]{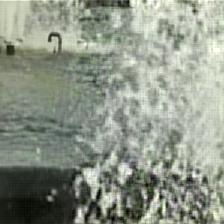}
\includegraphics[width=.076\linewidth]{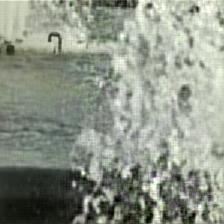}
\includegraphics[width=.076\linewidth]{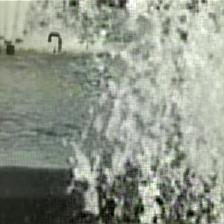}
\includegraphics[width=.076\linewidth]{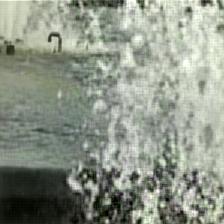}
(g) waterfall in a mountain \hspace{58mm} (h) water spray in a fountain\\[3px]
\caption{Synthesizing dynamic textures with only temporal stationarity. For each category, the first row displays 6 frames of the observed sequence, and the second and third rows display the corresponding frames of two synthesized sequences generated by the learning algorithm. The observed and the synthesized videos are of the size 224 $\times$ 224 pixels $\times$ 70 frames. (a) flashing lights. (b) water spray. (c) fountain. (d) spring water. (e) burning fire heating a pot. (f) burning fire in a stove. (g) waterfall in a mountain. (h) water spray in a fountain. }
	\label{fig:DTresults}
\end{center}
\end{figure*}

\begin{figure}[h]
\begin{center}
\includegraphics[width=.15\linewidth]{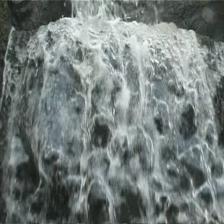}
\includegraphics[width=.15\linewidth]{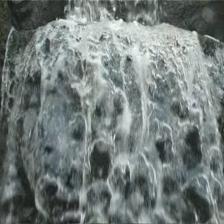}
\includegraphics[width=.15\linewidth]{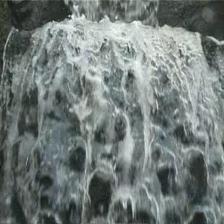}
\includegraphics[width=.15\linewidth]{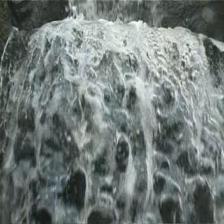}
\includegraphics[width=.15\linewidth]{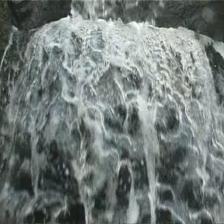}
\includegraphics[width=.15\linewidth]{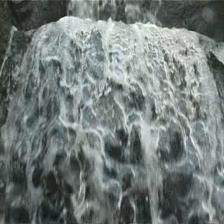} \\[3px]	

\includegraphics[width=.15\linewidth]{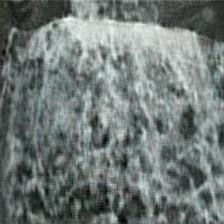}
\includegraphics[width=.15\linewidth]{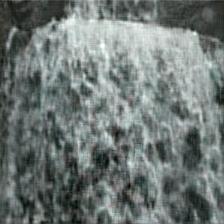}
\includegraphics[width=.15\linewidth]{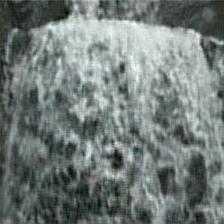}
\includegraphics[width=.15\linewidth]{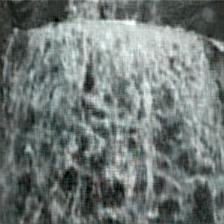}	
\includegraphics[width=.15\linewidth]{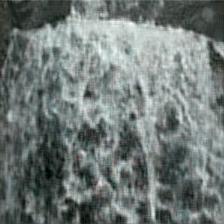}
\includegraphics[width=.15\linewidth]{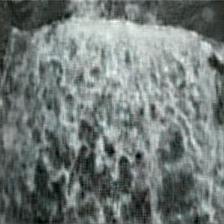}	\\[3px]

\includegraphics[width=.15\linewidth]{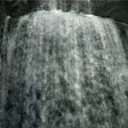}
\includegraphics[width=.15\linewidth]{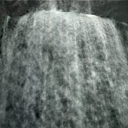}
\includegraphics[width=.15\linewidth]{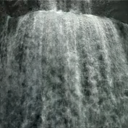}
\includegraphics[width=.15\linewidth]{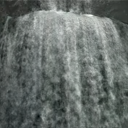}
\includegraphics[width=.15\linewidth]{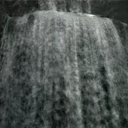}
\includegraphics[width=.15\linewidth]{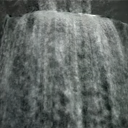} \\[3px]
	\caption{Comparison on synthesizing dynamic texture of waterfall. From top to bottom: segments of the observed sequence, synthesized sequence by our method, and synthesized sequence by the
method of \cite{doretto2003dynamic}.}
	\label{fig:DT_comp}
\end{center}
\end{figure}

We learn an energy-based spatial-temporal generative ConvNet for each category from one training video. We synthesize $\tilde{M}=3$ videos using the Langevin dynamics. Figure \ref{fig:DTresults} displays the results. For each category, the first row shows 6 frames of the observed sequence (224 $\times$ 224 pixels $\times$ 70 frames), and the second and third rows show the corresponding frames of two synthesized sequences generated by the learning algorithm. We use the same set of parameters for all the categories without tuning. Figure \ref{fig:DT_comp} compares our method to that of \cite{doretto2003dynamic}, which is a linear dynamic system model. The image sequence generated by this model appears more blurred than the sequence generated by our method.

The computational time for each iteration including 20 steps of Langevin dynamics is roughly 120 seconds on a PC with an Intel  i7-6700k CPU and a Titan Xp GPU. It takes roughly 5 seconds for each step of Langevin dynamics sampling.  

Quantitative evaluation for dynamic texture synthesis is a particularly challenging task, because there is no unique correct output
when synthesizing new samples of an observed dynamic texture. The structural similarity (SSIM) index \cite{wang2004image} is designed to provide a perceptual judgment on the similarity between two videos, and ranges from -1 to 1, with a larger score indicating greater similarity. We can calculate the SSIM between the synthesized image sequence and the original image sequence to examine the synthesis quality. A larger SSIM indicates a better synthesis quality due to higher perceptual similarity between the synthesized and original sequences. We compare our method with some baseline methods for dynamic textures, such as HOSVD \cite{costantini2008higher}, FFT-LDS \cite{abraham2005dynamic}, and Doretto et al \cite{doretto2003dynamic}, in terms of SSIM on 17 dynamic texture videos in Table \ref{compExp}. Our method outperforms the baseline methods. 

\begin{table}
\caption{A comparison of different dynamic texture models on SSIM}\label{compExp}
\begin{center}
\begin{tabular}{|l|c|c|c|c|}
\hline
& ours    & HOSVD   &  FFT-LDS  & Doretto\\ 
&     & \cite{costantini2008higher}    &  \cite{abraham2005dynamic} & \cite{doretto2003dynamic} \\ \hline \hline
boiling water &\textbf{0.8890}  & 0.4777  & 0.4719 &  0.8628 \\ 
falling water &\textbf{0.4044}  & 0.2246  & 0.2421 &  0.3458 \\  
fire flames&\textbf{0.4893}  & 0.2180  & 0.1491 &  0.4086 \\  
fire in a stove &\textbf{0.9259}  & 0.4646  & 0.4895 &  0.9075 \\ 
fire heating a pot &\textbf{0.7969}  & 0.4350  & 0.4401 &  0.7853 \\  
flashing lights &\textbf{0.7960}  & 0.2368  & 0.1982 &  0.7561 \\  
fountain & \textbf{0.3487}  & 0.1572  & 0.1114 &  0.3128 \\  
fountain spray & \textbf{0.5422}  & 0.2366  & 0.1948 &  0.4981 \\  
rocky waterfall &\textbf{0.6245}  & 0.1919  & 0.2281 &  0.6183 \\  
round fountain & \textbf{0.6662}  & 0.2082  & 0.2160 &  0.6617 \\  
spring water & \textbf{0.5703}  & 0.1587 & 0.2771 &  0.5561 \\  
washing machine & \textbf{0.9175}  & 0.4520  & 0.6026 &  0.9032 \\ 
water spray & \textbf{0.3528}  & 0.1615  & 0.0937 &  0.3365 \\   
water spray in a pool & \textbf{0.5610}  & 0.2171  & 0.1560 &  0.3442 \\  
water stream &\textbf{0.5030}  & 0.1667  & 0.1808 &  0.4492 \\  
waterfall &\textbf{0.3045}  & 0.1473  & 0.1001 &  0.2566 \\ 
waterfall in a mountain &\textbf{0.5415}  & 0.2272  & 0.2156 &  0.5206 \\ \hline \hline
Avg. & \textbf{0.6020}  & 0.2577  & 0.2569 & 0.5602 \\ \hline
\end{tabular}\\
\vskip 0.08in
\end{center}
\end{table}

Human perception, which is also an important and reliable measure to evaluate synthesis quality, has been used in \cite{chen2017photographic}, \cite{tesfaldet2018two}, \cite{wang2018high},\cite{XieGaoZhengZhuWu2019}. Following the protocal of\cite{XieGaoZhengZhuWu2019}, we conduct a human perceptual study to evaluate the perceived realism of the synthesized dynamic textures. Twenty different human observers are randomly chosen to participate in the perceptual experiment. Each participant performs 36 (12 categories $\times$ 3 examples) pairwise comparisons between a synthesized dynamic texture and its real version. Participants are asked to select which one looks more realistic after viewing each pair of dynamic textures for a specified exposure time. The dynamic textures are all shown at the same resolution (each image frame is of the size $64 \times 64$ pixels) in the form of video. The comparisons are randomized across the left-right order of two videos in each pair and the order of presenting different video pairs. The exposure time is chosen from discrete durations between 0.3 and 3.6 seconds. This evaluates how quickly the difference between two dynamic textures can be perceived.

We record the ``fooling'' rate, which is the participant error rate in distinguishing  synthesized dynamic textures from real ones, for each participant. The higher the ``fooling'' rate, the more realistic the synthesized dynamic textures. Note that ``perfectly'' synthesized results would lead to a ``fooling'' rate of 50$\%$ , indicating that the participants are unable to differentiate between the synthesized and real examples, and thus make random guesses.

Our model is compared with four baseline methods, such as LDS (linear dynamic system) \cite{doretto2003dynamic}, dynamic generator \cite{XieGaoZhengZhuWu2019}, TwoStream \cite{tesfaldet2018two} and MoCoGAN \cite{tulyakov2018mocogan}, on 12 dynamic texture video sequences (e.g., waterfall, burning fire, waving flag, etc) that have been adopted in \cite{XieGaoZhengZhuWu2019}. We briefly introduce them as follows. LDS simulates dynamic textures by a first-order auto-regressive moving average model driven by white zero-mean Gaussian noise; Dynamic generator is a non-linear generalization of the linear state space model where the non-linear transformations in the transition and emission models are parameterized by neural networks, and its training is based on maximum likelihood, which is accomplished by the alternating back-propagation through time algorithm; TwoStream method synthesizes dynamic textures by matching the feature statistics that are extracted from two pre-trained convolutional neural networks between synthesized and observed examples, where the pre-trained convolutional neural networks are trained for two independent tasks: object recognition, and optical flow prediction; and MoCoGAN is a motion and content decomposed generative adversarial network for video generation, where extra networks, e.g., image and video discriminators, are recruited to train the video generator in an adversarial manner. 

\begin{figure}[h]
\begin{center}
\includegraphics[width=.94\linewidth]{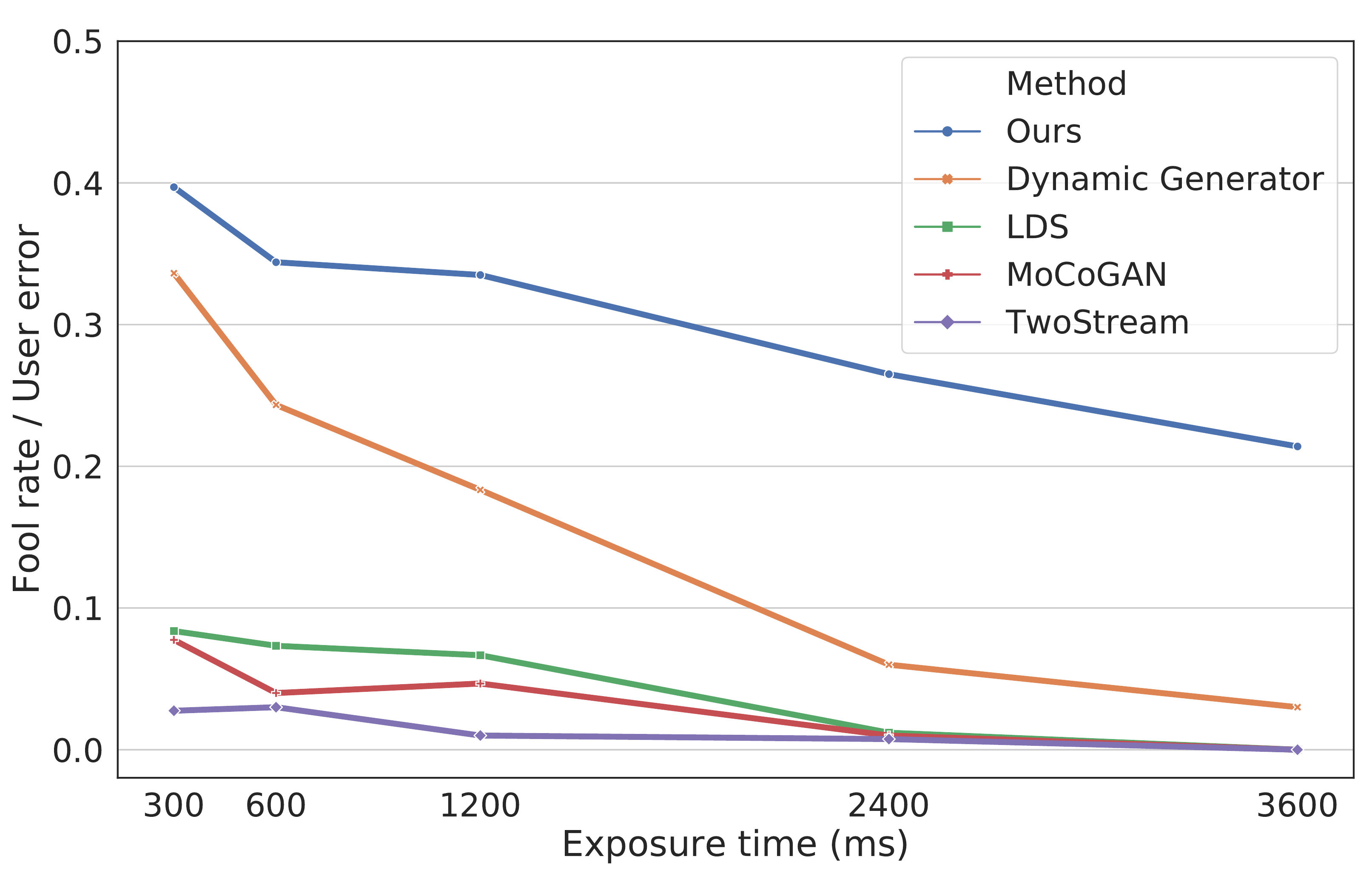}
	\caption{Limited time pairwise comparison results. Each curve shows the ``fooling'' rate / user error (realism) over different exposure times. The number of pairwise comparisons is 36. The number of participants is 20. }
	\label{fig:human_study}
\end{center}
\end{figure}

Figure \ref{fig:human_study} shows the comparison results at different exposure times, where the ``fooling'' rate decreases as exposure time increases, and then remains at the same level for longer exposures. This means that as the given exposure time becomes longer and longer, it becomes easier for the participants to tell ``fake'' examples from real ones. In summary, the dynamic textures generated by our proposed method are more realistic than other baseline methods. Besides, since the dynamic generator is a non-linear generalization of LDS, it is undoubtedly more expressive than LDS. The result also shows that the LDS outperforms those methods using sophisticated deep network structures (i.e., TwoStream and MoCoGAN). This is not surprising because when learning from a small number of training data (in this experiment, we only have one single training example), the MoCoGAN, with a large number of parameters that need to be estimated, may not fit the training data very well by using an unstable and complicated adversarial training scheme, while the TwoStream method has a limitation that it cannot synthesize dynamic textures not being  spatially homogeneous (i.e., dynamic textures with structured background, e.g., burning fire heating a pot). Our model is simple yet powerful, relying on neither extra networks nor extra labeled data for pre-training.

\begin{figure}[h]
\begin{center}
\includegraphics[width=.09\linewidth]{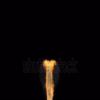}
\includegraphics[width=.09\linewidth]{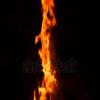}
\includegraphics[width=.09\linewidth]{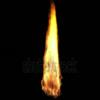}
\includegraphics[width=.09\linewidth]{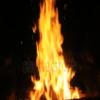}
\includegraphics[width=.09\linewidth]{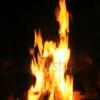}
\includegraphics[width=.09\linewidth]{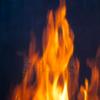}
\includegraphics[width=.09\linewidth]{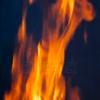}
\includegraphics[width=.09\linewidth]{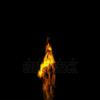}
\includegraphics[width=.09\linewidth]{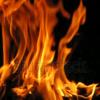}
\includegraphics[width=.09\linewidth]{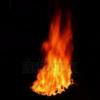} \\ \vspace{2pt}
\includegraphics[width=.09\linewidth]{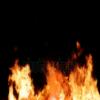}
\includegraphics[width=.09\linewidth]{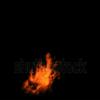}
\includegraphics[width=.09\linewidth]{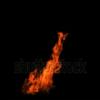}
\includegraphics[width=.09\linewidth]{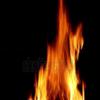}
\includegraphics[width=.09\linewidth]{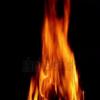}
\includegraphics[width=.09\linewidth]{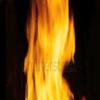}
\includegraphics[width=.09\linewidth]{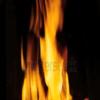}
\includegraphics[width=.09\linewidth]{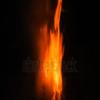}
\includegraphics[width=.09\linewidth]{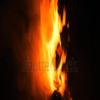}
\includegraphics[width=.09\linewidth]{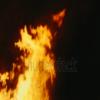} \\ \vspace{2pt}
\includegraphics[width=.09\linewidth]{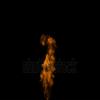}
\includegraphics[width=.09\linewidth]{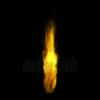}
\includegraphics[width=.09\linewidth]{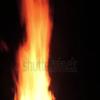}
\includegraphics[width=.09\linewidth]{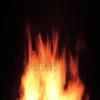}
\includegraphics[width=.09\linewidth]{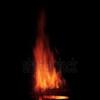}
\includegraphics[width=.09\linewidth]{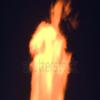}
\includegraphics[width=.09\linewidth]{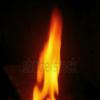}
\includegraphics[width=.09\linewidth]{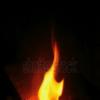}
\includegraphics[width=.09\linewidth]{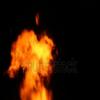}
\includegraphics[width=.09\linewidth]{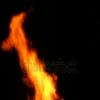}\\  \vspace{-3pt} {\footnotesize (a) $21$-st frame of 30 observed sequences} \\ \vspace{8pt}

\includegraphics[width=.09\linewidth]{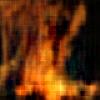}
\includegraphics[width=.09\linewidth]{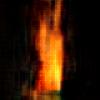}
\includegraphics[width=.09\linewidth]{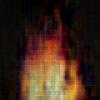}
\includegraphics[width=.09\linewidth]{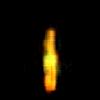}
\includegraphics[width=.09\linewidth]{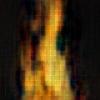}
\includegraphics[width=.09\linewidth]{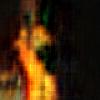}
\includegraphics[width=.09\linewidth]{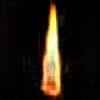}
\includegraphics[width=.09\linewidth]{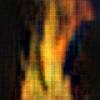}
\includegraphics[width=.09\linewidth]{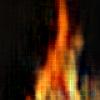}
\includegraphics[width=.09\linewidth]{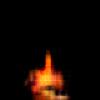}\\ \vspace{2pt}
\includegraphics[width=.09\linewidth]{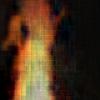}
\includegraphics[width=.09\linewidth]{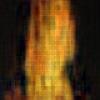}
\includegraphics[width=.09\linewidth]{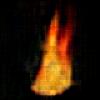}
\includegraphics[width=.09\linewidth]{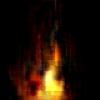}
\includegraphics[width=.09\linewidth]{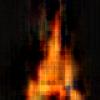}
\includegraphics[width=.09\linewidth]{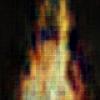}
\includegraphics[width=.09\linewidth]{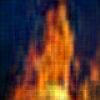}
\includegraphics[width=.09\linewidth]{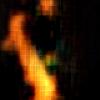}
\includegraphics[width=.09\linewidth]{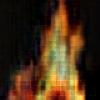}
\includegraphics[width=.09\linewidth]{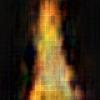}\\ \vspace{2pt}
\includegraphics[width=.09\linewidth]{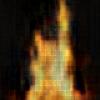}
\includegraphics[width=.09\linewidth]{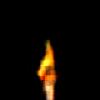}
\includegraphics[width=.09\linewidth]{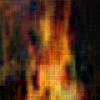}
\includegraphics[width=.09\linewidth]{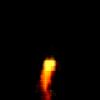}
\includegraphics[width=.09\linewidth]{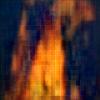}
\includegraphics[width=.09\linewidth]{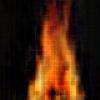}
\includegraphics[width=.09\linewidth]{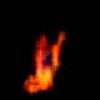}
\includegraphics[width=.09\linewidth]{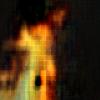}
\includegraphics[width=.09\linewidth]{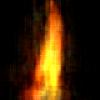}
\includegraphics[width=.09\linewidth]{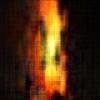}\\  \vspace{-3pt} {\footnotesize (b) $21$-st frame of 30 synthesized sequences} \\ \vspace{8pt}

\includegraphics[width=.09\linewidth]{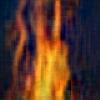}
\includegraphics[width=.09\linewidth]{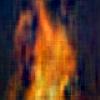}
\includegraphics[width=.09\linewidth]{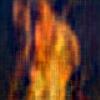}
\includegraphics[width=.09\linewidth]{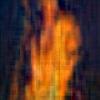}
\includegraphics[width=.09\linewidth]{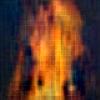}
\includegraphics[width=.09\linewidth]{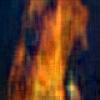}
\includegraphics[width=.09\linewidth]{./big_data/flame/iter_1350/synthesis_32/image_0021.jpg}
\includegraphics[width=.09\linewidth]{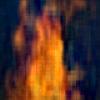}
\includegraphics[width=.09\linewidth]{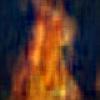}
\includegraphics[width=.09\linewidth]{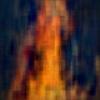}\\ \vspace{3pt}

\includegraphics[width=.09\linewidth]{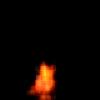}
\includegraphics[width=.09\linewidth]{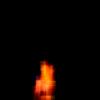}
\includegraphics[width=.09\linewidth]{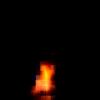}
\includegraphics[width=.09\linewidth]{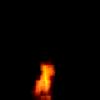}
\includegraphics[width=.09\linewidth]{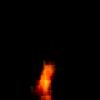}
\includegraphics[width=.09\linewidth]{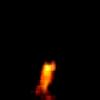}
\includegraphics[width=.09\linewidth]{./big_data/flame/iter_1350/synthesis_31/image_0021.jpg}
\includegraphics[width=.09\linewidth]{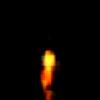}
\includegraphics[width=.09\linewidth]{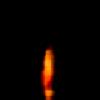}
\includegraphics[width=.09\linewidth]{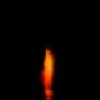}\\  \vspace{3pt} 

\includegraphics[width=.09\linewidth]{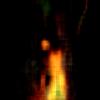}
\includegraphics[width=.09\linewidth]{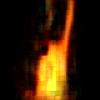}
\includegraphics[width=.09\linewidth]{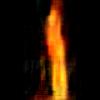}
\includegraphics[width=.09\linewidth]{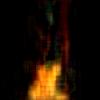}
\includegraphics[width=.09\linewidth]{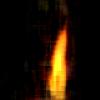}
\includegraphics[width=.09\linewidth]{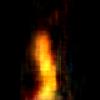}
\includegraphics[width=.09\linewidth]{./big_data/flame/iter_1350/synthesis_38/image_0021.jpg}
\includegraphics[width=.09\linewidth]{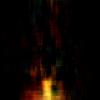}
\includegraphics[width=.09\linewidth]{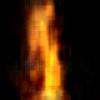}
\includegraphics[width=.09\linewidth]{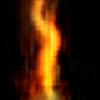}\\  \vspace{-3pt}
{\footnotesize (c) 3 examples of synthesized sequences} \\ 
	\caption{Learning  from 30 observed burning fire videos with mini-batch implementation. The batch size is 10. For each mini-batch, the number of parallel chains for synthesis is 13. The observed and synthesized videos are of the size 100 $\times $100 pixels $\times$ 30 frames. (a) displays one frame for each of 30 observed sequences. (b) displays the corresponding frame for each of the synthesized sequences. (c) shows three examples (one example per row) of synthesized video sequences. }
	\label{fig:flames}
\end{center}
\end{figure}

\begin{figure}[h]
\begin{center}
\includegraphics[width=.09\linewidth]{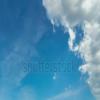}
\includegraphics[width=.09\linewidth]{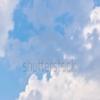}
\includegraphics[width=.09\linewidth]{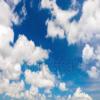}
\includegraphics[width=.09\linewidth]{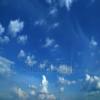}
\includegraphics[width=.09\linewidth]{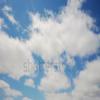}
\includegraphics[width=.09\linewidth]{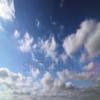}
\includegraphics[width=.09\linewidth]{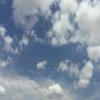}
\includegraphics[width=.09\linewidth]{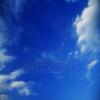}
\includegraphics[width=.09\linewidth]{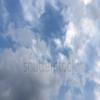}
\includegraphics[width=.09\linewidth]{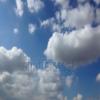} \\ \vspace{2pt}
\includegraphics[width=.09\linewidth]{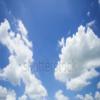}
\includegraphics[width=.09\linewidth]{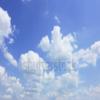}
\includegraphics[width=.09\linewidth]{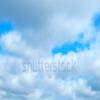}
\includegraphics[width=.09\linewidth]{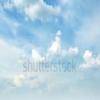}
\includegraphics[width=.09\linewidth]{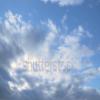}
\includegraphics[width=.09\linewidth]{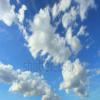}
\includegraphics[width=.09\linewidth]{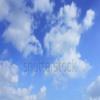}
\includegraphics[width=.09\linewidth]{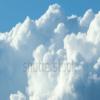}
\includegraphics[width=.09\linewidth]{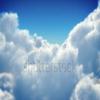}
\includegraphics[width=.09\linewidth]{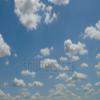} \\ \vspace{2pt}
\includegraphics[width=.09\linewidth]{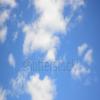}
\includegraphics[width=.09\linewidth]{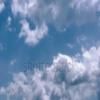}
\includegraphics[width=.09\linewidth]{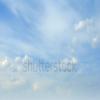}
\includegraphics[width=.09\linewidth]{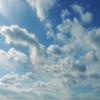}
\includegraphics[width=.09\linewidth]{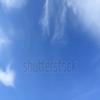}
\includegraphics[width=.09\linewidth]{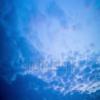}
\includegraphics[width=.09\linewidth]{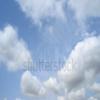}
\includegraphics[width=.09\linewidth]{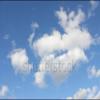}
\includegraphics[width=.09\linewidth]{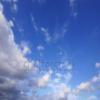}
\includegraphics[width=.09\linewidth]{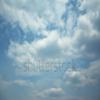}\\  \vspace{-3pt} {\footnotesize (a) $21$-st frame of 30 observed sequences} \\ \vspace{8pt}
\includegraphics[width=.09\linewidth]{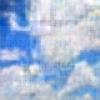}
\includegraphics[width=.09\linewidth]{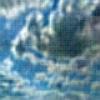}
\includegraphics[width=.09\linewidth]{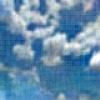}
\includegraphics[width=.09\linewidth]{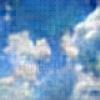}
\includegraphics[width=.09\linewidth]{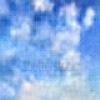}
\includegraphics[width=.09\linewidth]{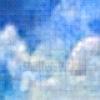}
\includegraphics[width=.09\linewidth]{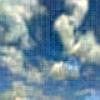}
\includegraphics[width=.09\linewidth]{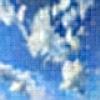}
\includegraphics[width=.09\linewidth]{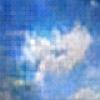}
\includegraphics[width=.09\linewidth]{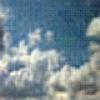}\\ \vspace{2pt}
\includegraphics[width=.09\linewidth]{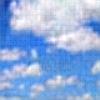}
\includegraphics[width=.09\linewidth]{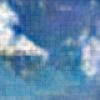}
\includegraphics[width=.09\linewidth]{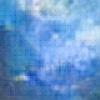}
\includegraphics[width=.09\linewidth]{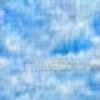}
\includegraphics[width=.09\linewidth]{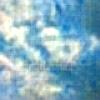}
\includegraphics[width=.09\linewidth]{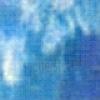}
\includegraphics[width=.09\linewidth]{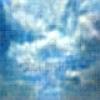}
\includegraphics[width=.09\linewidth]{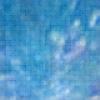}
\includegraphics[width=.09\linewidth]{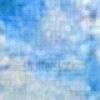}
\includegraphics[width=.09\linewidth]{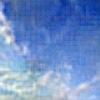}\\ \vspace{2pt}
\includegraphics[width=.09\linewidth]{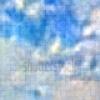}
\includegraphics[width=.09\linewidth]{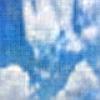}
\includegraphics[width=.09\linewidth]{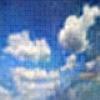}
\includegraphics[width=.09\linewidth]{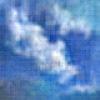}
\includegraphics[width=.09\linewidth]{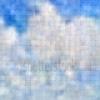}
\includegraphics[width=.09\linewidth]{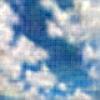}
\includegraphics[width=.09\linewidth]{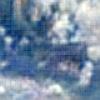}
\includegraphics[width=.09\linewidth]{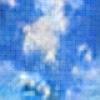}
\includegraphics[width=.09\linewidth]{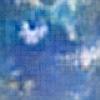}
\includegraphics[width=.09\linewidth]{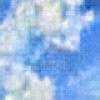}\\  \vspace{-3pt} {\footnotesize (b) $21$-st frame of 30 synthesized sequences} \\ \vspace{8pt}
\includegraphics[width=.09\linewidth]{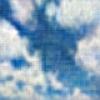}
\includegraphics[width=.09\linewidth]{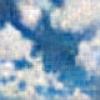}
\includegraphics[width=.09\linewidth]{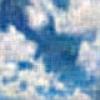}
\includegraphics[width=.09\linewidth]{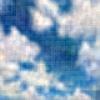}
\includegraphics[width=.09\linewidth]{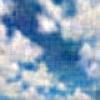}
\includegraphics[width=.09\linewidth]{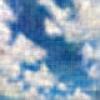}
\includegraphics[width=.09\linewidth]{./big_data/sky_cloud/iter_3000/synthesis_33/image_0021.jpg}
\includegraphics[width=.09\linewidth]{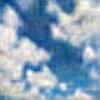}
\includegraphics[width=.09\linewidth]{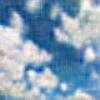}
\includegraphics[width=.09\linewidth]{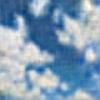}\\ \vspace{3pt}
\includegraphics[width=.09\linewidth]{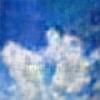}
\includegraphics[width=.09\linewidth]{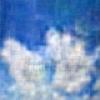}
\includegraphics[width=.09\linewidth]{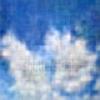}
\includegraphics[width=.09\linewidth]{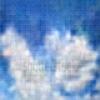}
\includegraphics[width=.09\linewidth]{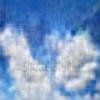}
\includegraphics[width=.09\linewidth]{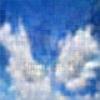}
\includegraphics[width=.09\linewidth]{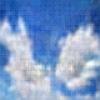}
\includegraphics[width=.09\linewidth]{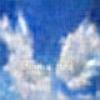}
\includegraphics[width=.09\linewidth]{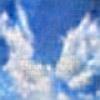}
\includegraphics[width=.09\linewidth]{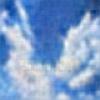}\\  \vspace{3pt}
\includegraphics[width=.09\linewidth]{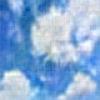}
\includegraphics[width=.09\linewidth]{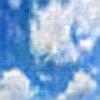}
\includegraphics[width=.09\linewidth]{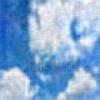}
\includegraphics[width=.09\linewidth]{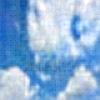}
\includegraphics[width=.09\linewidth]{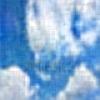}
\includegraphics[width=.09\linewidth]{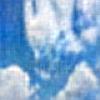}
\includegraphics[width=.09\linewidth]{./big_data/sky_cloud/iter_3000/synthesis_29/image_0021.jpg}
\includegraphics[width=.09\linewidth]{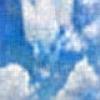}
\includegraphics[width=.09\linewidth]{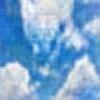}
\includegraphics[width=.09\linewidth]{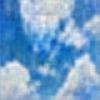}\\  \vspace{-3pt} {\footnotesize (c) 3 examples of synthesized sequences} \\ 
	\caption{Learning from 30 observed flowing cloud videos with mini-batch implementation.  The batch size is 10. For each mini-batch, the number of parallel chains for synthesis is 13. The observed and synthesized videos are of the size 100 $\times $100 pixels $\times$ 30 frames. (a) displays one frame for each of 30 observed sequences. (b) displays the corresponding frame for each of the synthesized sequences. (c) shows three examples (one example per row) of synthesized video sequences. }
	\label{fig:sky}
\end{center}
\end{figure}

The learning of our model can be scaled up. We learn the fire pattern from 30 training videos, with mini-batch implementation. The size of each mini-batch is 10 videos. Each video contains 30 frames ($100 \times 100$ pixels). For each mini-batch, ${\tilde M} = 13$ parallel chains for Langevin sampling is used. The gradient to update the model parameters is averaged over all mini-batches. For this experiment, we slightly modify the network by using $120$ $11 \times 11 \times 9$ filters with sub-sampling size of $5$ pixels and $4$ frames at the first layer, and 30 spatially fully connected filters with temporal size of 5 frames and sub-sampling size of $2$ at the second layer, while keeping the setting of the third layer unchanged. The number of learning iterations $T=1300$. Figure \ref{fig:flames} shows the result of learning fire patterns by displaying one frame for each of 30 observed sequences, the corresponding frame for each of the synthesized sequences, and three examples of synthesized sequences. Figure \ref{fig:sky} shows another example of learning flowing cloud patterns from 30 training videos. 

\subsection{Experiment 3: Generating action patterns without spatial or temporal stationarity}

Experiments 1 and 2 show that the energy-based generative spatial-temporal ConvNet can learn from video sequences without alignment. We can also specialize it to learning roughly aligned video sequences of action patterns, which are non-stationary in either spatial or temporal domain, by using a single
top-layer filter that covers the whole video sequence. 

The action patterns are spatial-temporally  aligned in the sense that (1) for each time step, the target objects in different videos possess the same locations, shapes, and poses (i.e., spatially aligned), and (2) the start and end times of the actions in different videos are the same (i.e., temporally aligned). 

We learn a 2-layer energy-based spatial-temporal generative  ConvNet from video sequences of roughly aligned actions. The first layer has 200 $7 \times 7  \times 7$ filters with sub-sampling size of 3 pixels and frames. The second layer is a fully connected layer with a single filter that covers the whole sequence. The observed sequences are of the size $100 \times 200$ pixels $\times 70$ frames.

\begin{figure}[h]
\begin{center}
\includegraphics[width=.15\linewidth]{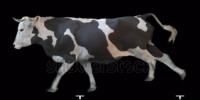}
\includegraphics[width=.15\linewidth]{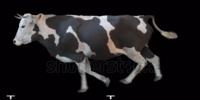}
\includegraphics[width=.15\linewidth]{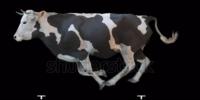}
\includegraphics[width=.15\linewidth]{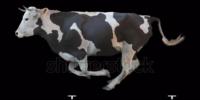}
\includegraphics[width=.15\linewidth]{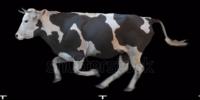}
\includegraphics[width=.15\linewidth]{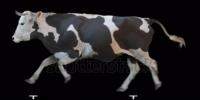} \\ \vspace{1pt}
\includegraphics[width=.15\linewidth]{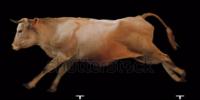}
\includegraphics[width=.15\linewidth]{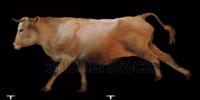}
\includegraphics[width=.15\linewidth]{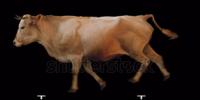}
\includegraphics[width=.15\linewidth]{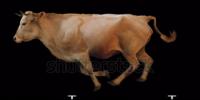}
\includegraphics[width=.15\linewidth]{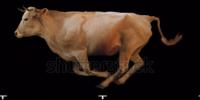}
\includegraphics[width=.15\linewidth]{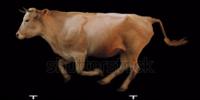}
\\ \vspace{-3pt} {\footnotesize observed sequences} \\ \vspace{3pt}

\includegraphics[width=.15\linewidth]{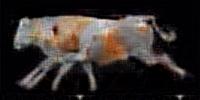}
\includegraphics[width=.15\linewidth]{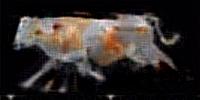}
\includegraphics[width=.15\linewidth]{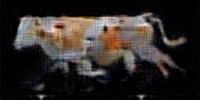}
\includegraphics[width=.15\linewidth]{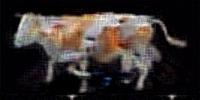}
\includegraphics[width=.15\linewidth]{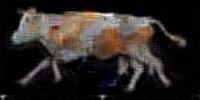}
\includegraphics[width=.15\linewidth]{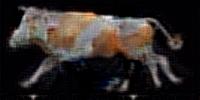} \\ \vspace{1pt}

\includegraphics[width=.15\linewidth]{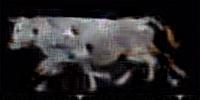}
\includegraphics[width=.15\linewidth]{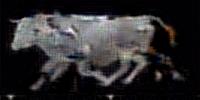}
\includegraphics[width=.15\linewidth]{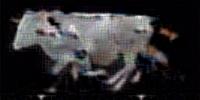}
\includegraphics[width=.15\linewidth]{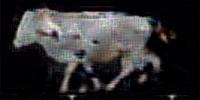}
\includegraphics[width=.15\linewidth]{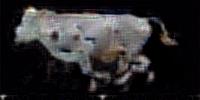}
\includegraphics[width=.15\linewidth]{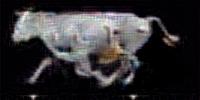}
\\ \vspace{-3pt} {\footnotesize synthesized sequences} \\ \vspace{2pt} (a) running cows \vspace{5pt}\\

\includegraphics[width=.15\linewidth]{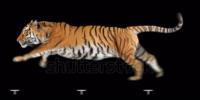}
\includegraphics[width=.15\linewidth]{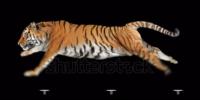}
\includegraphics[width=.15\linewidth]{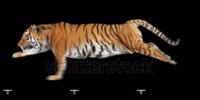}
\includegraphics[width=.15\linewidth]{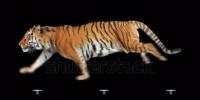}
\includegraphics[width=.15\linewidth]{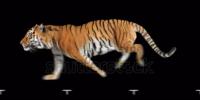}
\includegraphics[width=.15\linewidth]{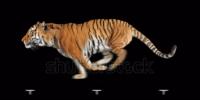} \\ \vspace{1pt}

\includegraphics[width=.15\linewidth]{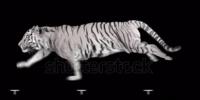}
\includegraphics[width=.15\linewidth]{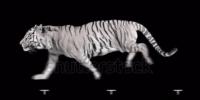}
\includegraphics[width=.15\linewidth]{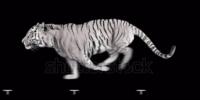}
\includegraphics[width=.15\linewidth]{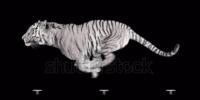}
\includegraphics[width=.15\linewidth]{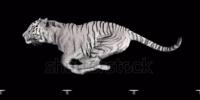}
\includegraphics[width=.15\linewidth]{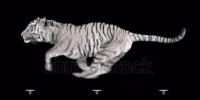}\\  \vspace{-3pt} {\footnotesize observed sequences} \vspace{3pt}

\includegraphics[width=.15\linewidth]{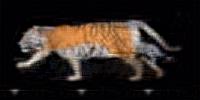}
\includegraphics[width=.15\linewidth]{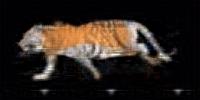}
\includegraphics[width=.15\linewidth]{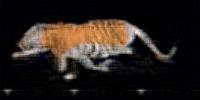}
\includegraphics[width=.15\linewidth]{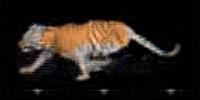}
\includegraphics[width=.15\linewidth]{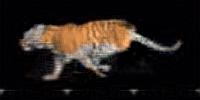}
\includegraphics[width=.15\linewidth]{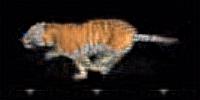} \\ \vspace{1pt}

\includegraphics[width=.15\linewidth]{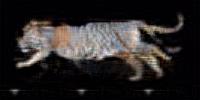}
\includegraphics[width=.15\linewidth]{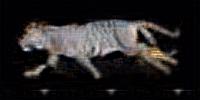}
\includegraphics[width=.15\linewidth]{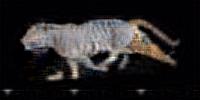}
\includegraphics[width=.15\linewidth]{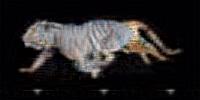}
\includegraphics[width=.15\linewidth]{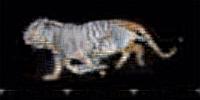}
\includegraphics[width=.15\linewidth]{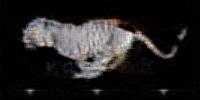}
\\  \vspace{-3pt} {\footnotesize synthesized sequences} \\  \vspace{2pt}(b) running tigers \\[3pt]

\includegraphics[width=.15\linewidth]{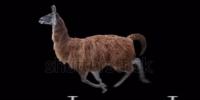}
\includegraphics[width=.15\linewidth]{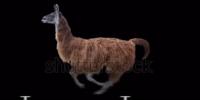}
\includegraphics[width=.15\linewidth]{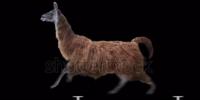}
\includegraphics[width=.15\linewidth]{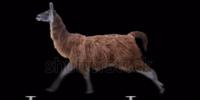}
\includegraphics[width=.15\linewidth]{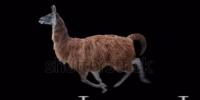}
\includegraphics[width=.15\linewidth]{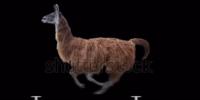} \\ \vspace{1pt}
\includegraphics[width=.15\linewidth]{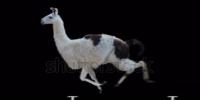}
\includegraphics[width=.15\linewidth]{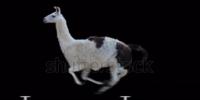}
\includegraphics[width=.15\linewidth]{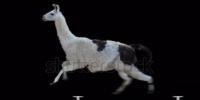}
\includegraphics[width=.15\linewidth]{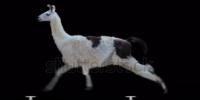}
\includegraphics[width=.15\linewidth]{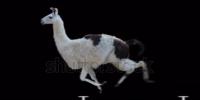}
\includegraphics[width=.15\linewidth]{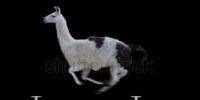}\\  \vspace{-3pt} {\footnotesize observed sequences} \vspace{3pt}

\includegraphics[width=.15\linewidth]{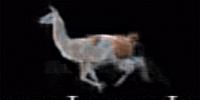}
\includegraphics[width=.15\linewidth]{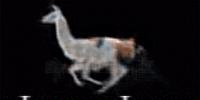}
\includegraphics[width=.15\linewidth]{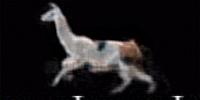}
\includegraphics[width=.15\linewidth]{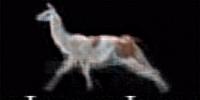}
\includegraphics[width=.15\linewidth]{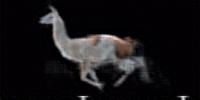}
\includegraphics[width=.15\linewidth]{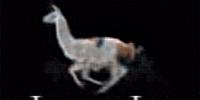} \\ \vspace{1pt}

\includegraphics[width=.15\linewidth]{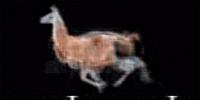}
\includegraphics[width=.15\linewidth]{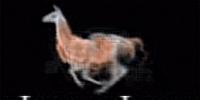}
\includegraphics[width=.15\linewidth]{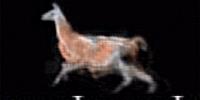}
\includegraphics[width=.15\linewidth]{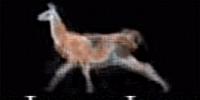}
\includegraphics[width=.15\linewidth]{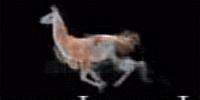}
\includegraphics[width=.15\linewidth]{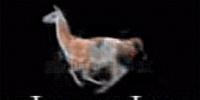}
\\  \vspace{-3pt} {\footnotesize synthesized sequences} \\  \vspace{2pt}(c) running llamas 
	\caption{Synthesizing action patterns without spatial or temporal stationarity. For each action video sequence, 6 continuous frames are shown. (a) running cows. Frames of 2 of 5 training sequences are displayed. The corresponding frames of 2 of 8 synthesized sequences generated by the learning algorithm are displayed. (b) running tigers. Frames of 2 observed training sequences are displayed. The corresponding frames of 2 of 4 synthesized sequences are displayed. (c) running llamas. Frames of 2 observed training sequences are displayed. The corresponding frames of 2 of 5 synthesized sequences are displayed. The observed sequences are of the size $100 \times  100$ pixels $\times 70$ frames.}
	\label{fig:action}
\end{center}
\end{figure}

Figure \ref{fig:action} displays three results of modeling and synthesizing actions from roughly  aligned video sequences. We learn a model for each category, where the number of training sequences  is 5 for the running cow example, 2 for the running tiger example, and 2 for the running llama example. The videos are collected from the Internet and each has 70 frames, with an animal running at the center of a black background. For each example, Figure \ref{fig:action} displays segments of 2 observed sequences, and segments of 2 synthesized action sequences generated by the learning algorithm. We run $\tilde{M}=8$ paralleled  chains for the experiment of running cows, 4 paralleled chains for the experiment of running tigers, and 5 paralleled chains for the experiments of running llamas. 

The experiments show that our model can capture non-stationary action patterns. In general, both appearances and motions of the animals in the synthesized video sequences are physically plausible. We also notice that the synthesized action patterns contain some artifacts (e.g., the appearance of additional limbs of the animals in the synthesized video sequences), which is due to the fact that the action patterns in the collected training video sequences are not perfectly aligned (e.g., at a specific time, the poses of the animals in different videos are not the same). Note that synthesizing action patterns can be difficult for the methods of \cite{doretto2003dynamic} and \cite{tesfaldet2018two}. 

One limitation of our model is that it does not involve explicit tracking of the objects and their parts, therefore the model can only learn descriptive features of an action pattern for the sake of synthesis, instead of an explainable action decomposition for the purpose of understanding. The other limitation is that a single model can only be used for single category of action pattern. Unsupervised learning from actions of mixed categories requires fitting a mixture of energy-based spatial-temporal generative ConvNet models. However, this would require the estimation of the intractable normalizing constants $Z(\theta)$ of the models, because the EM-type algorithm to fit mixtures of energy-based spatial-temporal generative ConvNet models involves computation of the log-likelihood of each video sequence under each model. Our previous works \cite{xie2014learning} and \cite{xie2015boosting} have studied this issue. Nevertheless, the dynamic generator \cite{XieGaoZhengZhuWu2019} and MoCoGAN \cite{tulyakov2018mocogan} adopt top-down  generators, with explicit latent variables accounting for action categories, to learn from actions of mixed categories in an unsupervised manner. 
	  
\subsection{Experiment 4: Learning from incomplete data}

Our model can learn from video sequences with occluded pixels. The task is inspired by the fact that most of the videos contain occluded objects. Our learning method can be adapted to this task with minimal modification. The modification involves, for each iteration, running $k$ steps of Langevin dynamics to recover the occluded regions of the observed sequences. The additional Langevin dynamics for recovery is initialized by the observed occluded sequences, and it only synthesizes the occluded regions, while keeping the visible parts in the observed sequences unchanged. At each iteration, we use the  completed observed sequences and the synthesized sequences to compute the gradient of the log-likelihood and update the model parameters. Our method simultaneously accomplishes the following tasks: (1) recover the occluded pixels of the training video sequences, (2) synthesize new video sequences from the learned model, (3) learn the model by updating the model parameters using the recovered sequences and the synthesized sequences. See Algorithm \ref{code:recovery} for the description of the learning, sampling, and recovery algorithm. 

\begin{algorithm}
\caption{Learning, sampling, and recovery algorithm}
\label{code:recovery}
\begin{algorithmic}[1]

\REQUIRE ~~\\
(1) training image sequences with occluded pixels $\{\I_m, m=1,...,M\}$ \\
(2) binary masks $\{O_m, m=1,...,M \}$ indicating the locations of the occluded pixels in the training image sequences\\  
(3) number of synthesized image sequences $\tilde{M}$\\
(4) number of Langevin steps $l$ for synthesizing image sequences\\
(5) number of Langevin steps $k$ for recovering the occluded pixels\\
(6) number of learning iterations $T$

\ENSURE~~\\
(1) estimated model parameters $\theta$\\
(2) synthesized image sequences $\{\tI_m, m = 1, ..., \tilde{M}\}$ \\
(3) recovered image sequences $\{\I^{'}_m, m=1,...,M\}$
\item[]
\STATE Let $t\leftarrow 0$, initialize $\theta^{(0)}$.
\STATE Initialize $\tI_m $, for $m = 1, ..., \tilde{M}$, by sampling from $q(\I)$. 
\STATE Initialize $\I^{'}_m = \I_m$, for $m = 1, ..., M$. 
\REPEAT 
\STATE For each $m$, run $k$ steps of Langevin dynamics to recover the occluded region of $\I^{'}_m$, i.e., starting from the current $\I^{'}_m$, each step 
follows equation (\ref{eq:Langevin}), but only the occluded pixels in $\I^{'}_m$ (specified by $O_m$) are updated in each step. 
\STATE For each $m$, run $l$ steps of Langevin dynamics to update $\tI_m$, i.e., starting from the current $\tI_m$, each step 
follows equation (\ref{eq:Langevin}). 
\STATE Calculate  $H^{\obs} = \sum_{m=1}^{M} \frac{\partial}{\partial \theta} f(\I^{'}_m; \theta^{(t)})/M$, and
$H^{\syn} =  \sum_{m=1}^{\tM} \frac{\partial}{\partial \theta} f(\tI_m; \theta^{(t)})/\tM$.
\STATE Update $\theta^{(t+1)} \leftarrow \theta^{(t)} + \eta ( H^{\obs} - H^{\syn}) $,  with step size $\eta$. 
\STATE Let $t \leftarrow t+1$
\UNTIL $t = T$
\end{algorithmic}
\end{algorithm}

\begin{table}[h]
\caption{Recovery errors in occlusion experiments}\label{recoveryExp}
\vskip 0.02in
\begin{center}
\begin{footnotesize}
(a) salt and pepper masks\\
\begin{tabular}{|c|c|c|c|}
\hline
           & ours   & MRF-$\ell_1$      & MRF-$\ell_2$  \\ \hline \hline
flag       & \textbf{3.7923} & 6.6211  & 10.9216  \\ \hline
fountain   & \textbf{5.5403} & 8.1904  & 11.3850  \\ \hline
ocean      & \textbf{3.3739} & 7.2983 & 9.6020     \\ \hline
playing    & \textbf{5.9035} & 14.3665& 15.7735    \\ \hline
sea world & \textbf{5.3720} &  10.6127  & 11.7803  \\ \hline
traffic    & \textbf{7.2029} & 14.7512 & 17.6790  \\ \hline
windmill   & \textbf{5.9484} &  8.9095   & 12.6487  \\ \hline \hline
Avg.   & \textbf{5.3048} &  10.1071    & 12.8272  \\ \hline
\end{tabular}
\vskip 0.08in
(b) single region masks \\
\begin{tabular}{|c|c|c|c|}
\hline
           & ours    & MRF-$\ell_1$      & MRF-$\ell_2$  \\ \hline \hline
flag       & \textbf{8.1636}  & 10.6586  & 12.5300\\ \hline
fountain   & \textbf{6.0323}  & 11.8299  & 12.1696\\ \hline
ocean      & \textbf{3.4842}  & 8.7498 & 9.8078\\ \hline
playing    & \textbf{6.1575}  & 15.6296& 15.7085\\ \hline
sea world & \textbf{5.8850} & 12.0297  & 12.2868\\ \hline
traffic    & \textbf{6.8306}  & 15.3660 & 16.5787\\ \hline
windmill   & \textbf{7.8858} & 11.7355   & 13.2036\\ \hline \hline
Avg.   & \textbf{ 6.3484} & 12.2856    & 13.1836\\ \hline
\end{tabular}
\vskip 0.08in
(c) 50$\%$ missing frames\\
\begin{tabular}{|c|c|c|c|}
\hline
           & ours    & MRF-$\ell_1$      & MRF-$\ell_2$  \\ \hline \hline
flag       & \textbf{5.5992} & 10.7171  & 12.6317\\ \hline
fountain   & \textbf{8.0531} & 19.4331  & 13.2251\\ \hline
ocean      & \textbf{4.0428}  & 9.0838 & 9.8913\\ \hline
playing    & \textbf{7.6103 } & 22.2827& 17.5692\\ \hline
sea world & \textbf{5.4348} & 13.5101  &12.9305\\ \hline
traffic    & \textbf{8.8245}  & 16.6965 & 17.1830\\ \hline
windmill   & \textbf{7.5346} & 13.3364  & 12.9911 \\ \hline \hline
Avg.   & \textbf{ 6.7285} & 15.0085   & 13.7746\\ \hline
\end{tabular}
\end{footnotesize}
\end{center}
\end{table}

\begin{figure*}[h]
\begin{center}

\includegraphics[width=.07\linewidth]{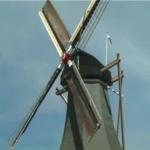}
\includegraphics[width=.07\linewidth]{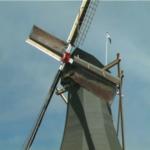}
\includegraphics[width=.07\linewidth]{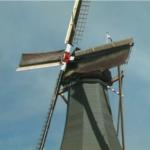}
\includegraphics[width=.07\linewidth]{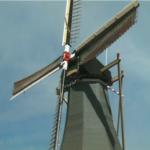}
\includegraphics[width=.07\linewidth]{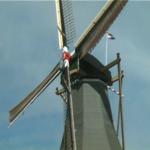}
\includegraphics[width=.07\linewidth]{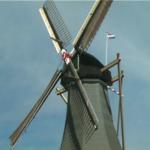} 
\hspace{7mm}
\includegraphics[width=.07\linewidth]{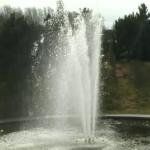}
\includegraphics[width=.07\linewidth]{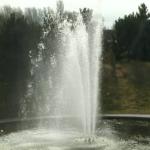}
\includegraphics[width=.07\linewidth]{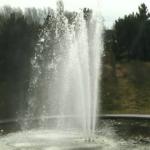}
\includegraphics[width=.07\linewidth]{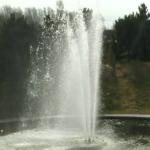}
\includegraphics[width=.07\linewidth]{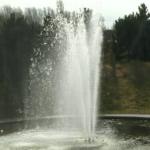}
\includegraphics[width=.07\linewidth]{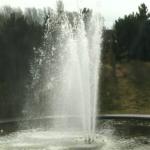} \\[3px]

\includegraphics[width=.07\linewidth]{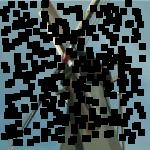}
\includegraphics[width=.07\linewidth]{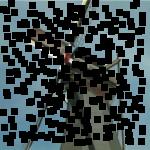}
\includegraphics[width=.07\linewidth]{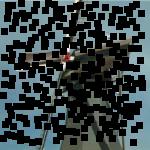}
\includegraphics[width=.07\linewidth]{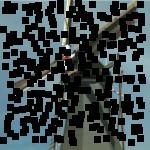}
\includegraphics[width=.07\linewidth]{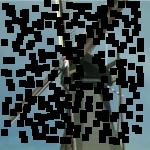}
\includegraphics[width=.07\linewidth]{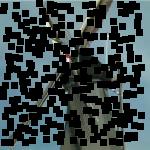} 
\hspace{7mm}
\includegraphics[width=.07\linewidth]{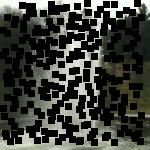}
\includegraphics[width=.07\linewidth]{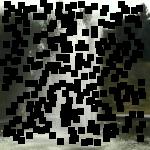}
\includegraphics[width=.07\linewidth]{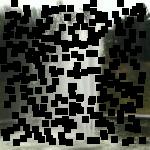}
\includegraphics[width=.07\linewidth]{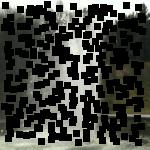}
\includegraphics[width=.07\linewidth]{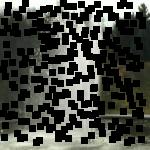}
\includegraphics[width=.07\linewidth]{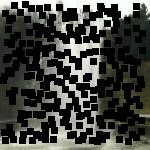} 
\\[3px]

\includegraphics[width=.07\linewidth]{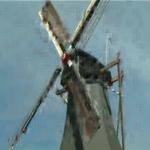}
\includegraphics[width=.07\linewidth]{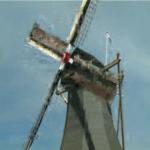}
\includegraphics[width=.07\linewidth]{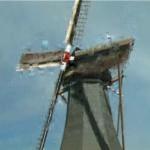}
\includegraphics[width=.07\linewidth]{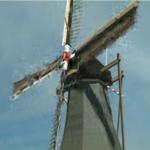}
\includegraphics[width=.07\linewidth]{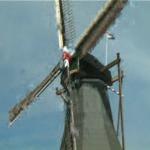}
\includegraphics[width=.07\linewidth]{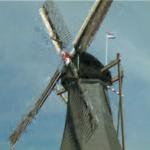}
\hspace{7mm}
\includegraphics[width=.07\linewidth]{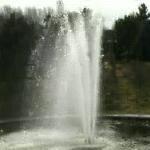}
\includegraphics[width=.07\linewidth]{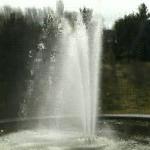}
\includegraphics[width=.07\linewidth]{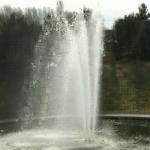}
\includegraphics[width=.07\linewidth]{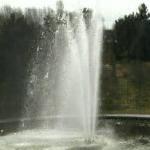}
\includegraphics[width=.07\linewidth]{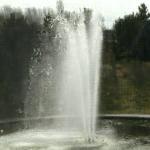}
\includegraphics[width=.07\linewidth]{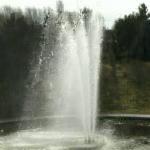}
\\ windmill \hspace{75mm} fountain
\\ (a) $50\%$ salt and pepper masks \\[8px]
\includegraphics[width=.07\linewidth]{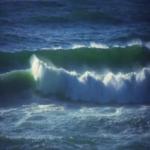}
\includegraphics[width=.07\linewidth]{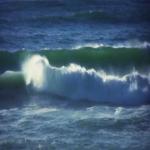}
\includegraphics[width=.07\linewidth]{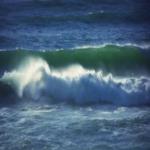}
\includegraphics[width=.07\linewidth]{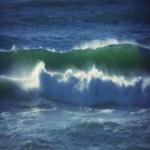}
\includegraphics[width=.07\linewidth]{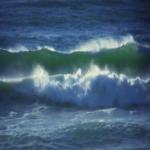}
\includegraphics[width=.07\linewidth]{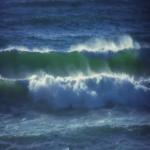} 
\hspace{7mm}
\includegraphics[width=.07\linewidth]{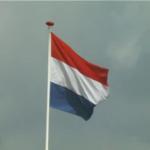}
\includegraphics[width=.07\linewidth]{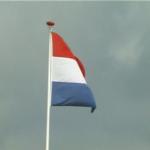}
\includegraphics[width=.07\linewidth]{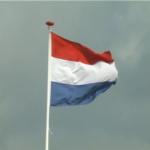}
\includegraphics[width=.07\linewidth]{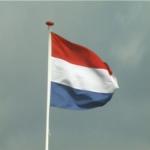}
\includegraphics[width=.07\linewidth]{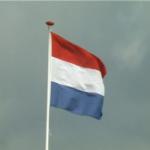}
\includegraphics[width=.07\linewidth]{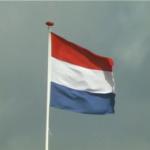} 
\\ [3px]
\includegraphics[width=.07\linewidth]{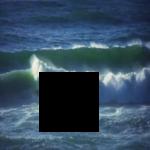}
\includegraphics[width=.07\linewidth]{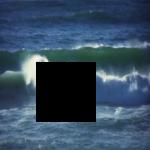}
\includegraphics[width=.07\linewidth]{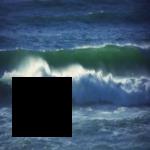}
\includegraphics[width=.07\linewidth]{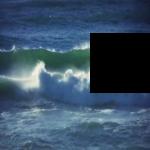}
\includegraphics[width=.07\linewidth]{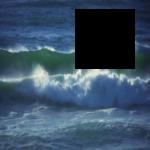}
\includegraphics[width=.07\linewidth]{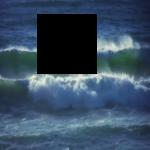} 
\hspace{7mm}
\includegraphics[width=.07\linewidth]{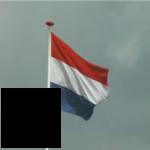}
\includegraphics[width=.07\linewidth]{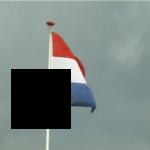}
\includegraphics[width=.07\linewidth]{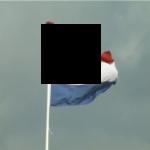}
\includegraphics[width=.07\linewidth]{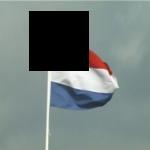}
\includegraphics[width=.07\linewidth]{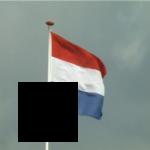}
\includegraphics[width=.07\linewidth]{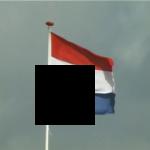} \\ [3px]

\includegraphics[width=.07\linewidth]{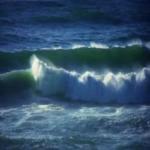}
\includegraphics[width=.07\linewidth]{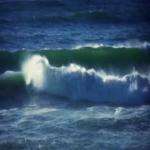}
\includegraphics[width=.07\linewidth]{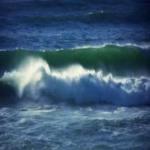}
\includegraphics[width=.07\linewidth]{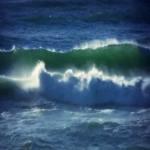}
\includegraphics[width=.07\linewidth]{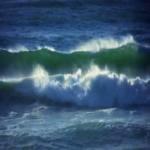}
\includegraphics[width=.07\linewidth]{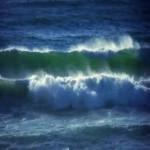}
\hspace{7mm}
\includegraphics[width=.07\linewidth]{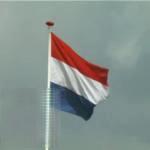}
\includegraphics[width=.07\linewidth]{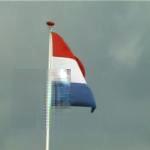}
\includegraphics[width=.07\linewidth]{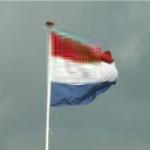}
\includegraphics[width=.07\linewidth]{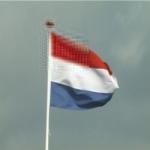}
\includegraphics[width=.07\linewidth]{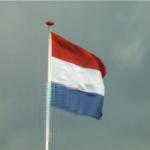}
\includegraphics[width=.07\linewidth]{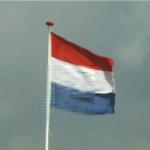}
\\ ocean \hspace{84mm} flag
\\ (b) single region masks \\[8px]

\includegraphics[width=.07\linewidth]{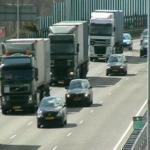}
\includegraphics[width=.07\linewidth]{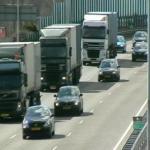}
\includegraphics[width=.07\linewidth]{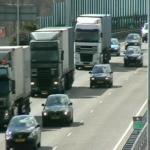}
\includegraphics[width=.07\linewidth]{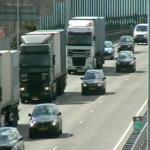}
\includegraphics[width=.07\linewidth]{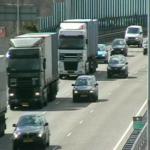}
\includegraphics[width=.07\linewidth]{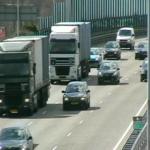}
\hspace{7mm}
\includegraphics[width=.07\linewidth]{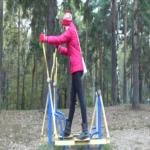}
\includegraphics[width=.07\linewidth]{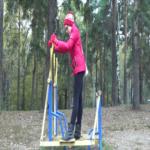}
\includegraphics[width=.07\linewidth]{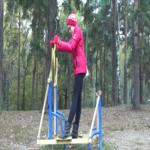}
\includegraphics[width=.07\linewidth]{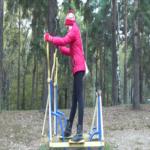}
\includegraphics[width=.07\linewidth]{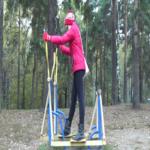}
\includegraphics[width=.07\linewidth]{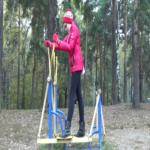}
\\[3px]

\includegraphics[width=.07\linewidth]{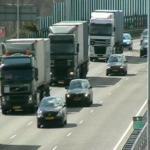}
\includegraphics[width=.07\linewidth]{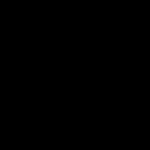}
\includegraphics[width=.07\linewidth]{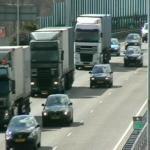}
\includegraphics[width=.07\linewidth]{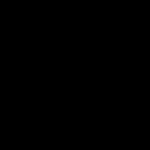}
\includegraphics[width=.07\linewidth]{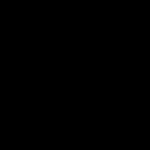}
\includegraphics[width=.07\linewidth]{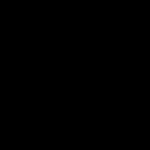}
\hspace{7mm}
\includegraphics[width=.07\linewidth]{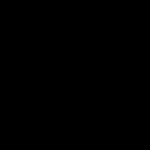}
\includegraphics[width=.07\linewidth]{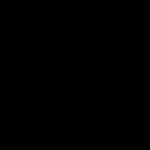}
\includegraphics[width=.07\linewidth]{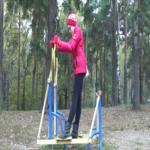}
\includegraphics[width=.07\linewidth]{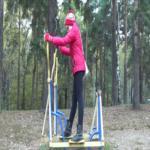}
\includegraphics[width=.07\linewidth]{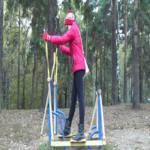}
\includegraphics[width=.07\linewidth]{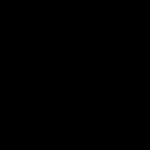}
\\[3px]
\includegraphics[width=.07\linewidth]{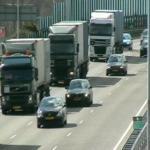}
\includegraphics[width=.07\linewidth]{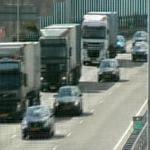}
\includegraphics[width=.07\linewidth]{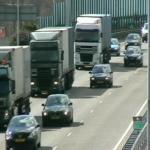}
\includegraphics[width=.07\linewidth]{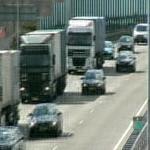}
\includegraphics[width=.07\linewidth]{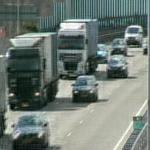}
\includegraphics[width=.07\linewidth]{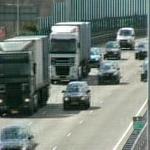}
\hspace{7mm}
\includegraphics[width=.07\linewidth]{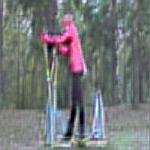}
\includegraphics[width=.07\linewidth]{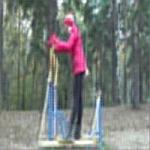}
\includegraphics[width=.07\linewidth]{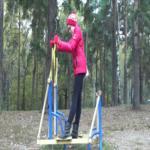}
\includegraphics[width=.07\linewidth]{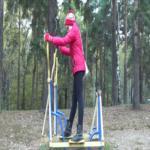}
\includegraphics[width=.07\linewidth]{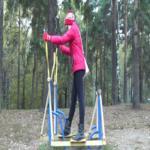}
\includegraphics[width=.07\linewidth]{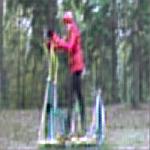}\\
traffic  \hspace{78mm} playing
\\ (c) $50\%$ missing frames \vspace{3pt}
\caption{Learning energy-based spatial-temporal generative ConvNets from occluded video sequences. For each experiment, the first row shows a segment of the observed sequence, the second row shows a segment of the occluded sequence with black masks, and the third row shows the corresponding segment of the recovered sequence. The observed occluded sequences are of the size $150 \times 150$ pixels $\times 70$ frames. (a) type 1: salt and pepper mask.  (b) type 2: single region mask. (c) type 3: $50\%$ missing frames. }
	\label{fig:recovery}
\end{center}
\end{figure*}

\begin{figure}[h]
\begin{center}
\includegraphics[width=.27\linewidth]{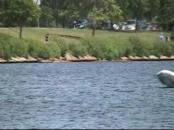}
\includegraphics[width=.27\linewidth]{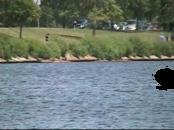}
\includegraphics[width=.27\linewidth]{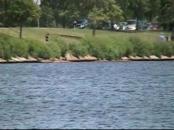}\\[3px]

\includegraphics[width=.27\linewidth]{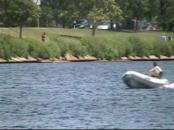}
\includegraphics[width=.27\linewidth]{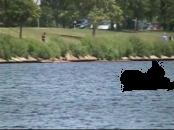}
\includegraphics[width=.27\linewidth]{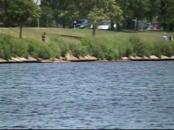}\\[3px]

\includegraphics[width=.27\linewidth]{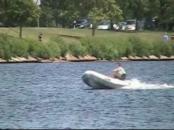}
\includegraphics[width=.27\linewidth]{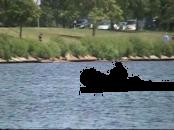}
\includegraphics[width=.27\linewidth]{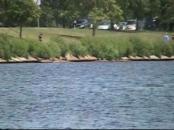}\\[3px]

\includegraphics[width=.27\linewidth]{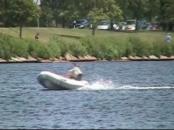}
\includegraphics[width=.27\linewidth]{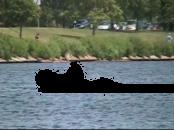}
\includegraphics[width=.27\linewidth]{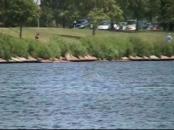}\\
(a) removing a moving boat in the lake\\[8px]

\includegraphics[width=.27\linewidth]{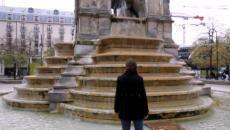}
\includegraphics[width=.27\linewidth]{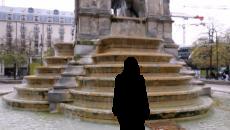}
\includegraphics[width=.27\linewidth]{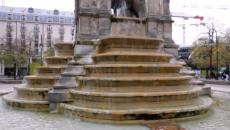}\\[3px]

\includegraphics[width=.27\linewidth]{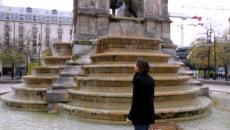}
\includegraphics[width=.27\linewidth]{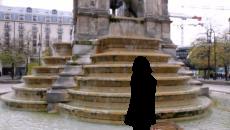}
\includegraphics[width=.27\linewidth]{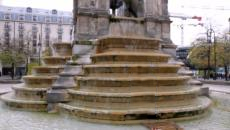}\\[3px]

\includegraphics[width=.27\linewidth]{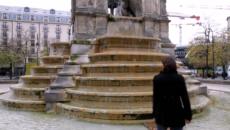}
\includegraphics[width=.27\linewidth]{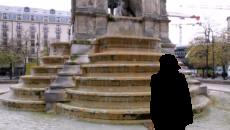}
\includegraphics[width=.27\linewidth]{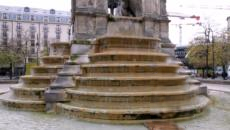}\\[3px]

\includegraphics[width=.27\linewidth]{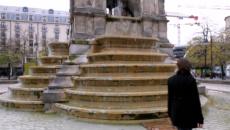}
\includegraphics[width=.27\linewidth]{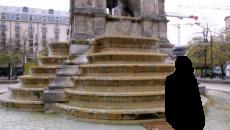}
\includegraphics[width=.27\linewidth]{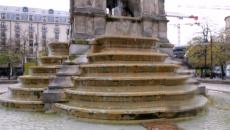}\\
(b) removing a walking person in front of fountain\\[3px]
	\caption{Background inpainting for videos. For each experiment, the first column displays 4 frames of the original video. The second column shows the corresponding frames with black masks occluding the target to be removed. The third column shows the inpainting result by our algorithm. (a) a moving boat in the lake ($130 \times 174$ pixels $\times 150$ frames). (b) a walking person in front of fountain ($130 \times 230$ pixels $\times 104$ frames).}
	\label{fig:bg_inpainting}
\end{center}
\end{figure}

We design 3 types of occlusions: (1) Type 1: salt and pepper occlusion, where we randomly place $7 \times 7$ masks on the $150 \times 150$ image domain to cover $50\%$ of the pixels of the videos. (2) Type 2: single region mask occlusion, where we randomly place a $60 \times 60$ mask on the $150 \times 150$ image domain. (3) Type 3: missing frames, where we randomly block $50\%$ of the image frames from each video. Figure \ref{fig:recovery} displays two examples of the recovery results for each type of occlusion. Each video has 70 frames. We can see that our model can recover the incomplete video sequences, while learning from them.   

To quantitatively evaluate the qualities of the recovered videos, we test our method on 7 video sequences, which are collected from DynTex++ dataset of \cite{ghanem2010maximum}, with 3 types of occlusions mentioned above. We use the same model structure as the one used in Experiment 3. The number of Langevin steps for recovery is set to be equal to the number of Langevin steps for synthesizing, which is 20. We run a single Langevin chain for synthesis. For each experiment, we report the recovery errors measured by the average per pixel difference between the original image sequence and the recovered image sequence on the occluded pixels. The range of pixel intensities is $[0,255]$. We compare our results with the results obtained by a baseline method, which is a generic Markov random field model defined on the video sequence. The model is a 3D (spatial-temporal) Markov random field, whose potentials are pairwise $\ell_1$ or $\ell_2$ differences between nearest neighbor pixels, where the nearest neighbors are defined in both the spatial and temporal domains. The image sequences are recovered by sampling the intensities of the occluded pixels conditional on the observed pixels using the Gibbs sampler. Table \ref{recoveryExp} shows the comparison results for 3 types of occlusions. Our model significantly outperforms the baseline methods in terms of recovery errors. Note that the recovery errors are not training errors, because the intensities of the occluded pixels of the image frames are not observed in training. 

We want to emphasize that (1) our models differ from denoising auto-encoders \cite{vincent2010stacked}, where the training data are fully observed, and noises are added as a matter of regularization; (2) our experiments are different from in-painting or de-noising, where the prior model or regularization has already been learned from fully observed data or provided; (3) Learning from incomplete data can be difficult for GAN and VAE.

\subsection{Experiment 5: Background inpainting}

If a moving object in the video is occluded in each frame, it turns out that the recovery algorithm will become an algorithm for background inpainting of videos, where the goal is to remove the undesired moving object from the video. We use the same model as the one in Experiment 2  for Figure \ref{fig:DTresults}. Figure \ref{fig:bg_inpainting} shows two examples of removals of (a) a moving boat and (b) a walking person  respectively. The videos are collected from \cite{data}. For each example, the first column displays 4 frames of the original video. The second column shows the corresponding frames with masks occluding the target to be removed. The third column presents the inpainting result by our algorithm. The video size is $130 \times 174$ pixels $\times 150$ frames in example (a) and $130 \times 230$ pixels $\times 104$ frames in example (b). The experiment is different from the video inpainting by interpolation. We synthesize image patches to fill in the empty regions of the video by running Langevin dynamics. We run a single Langevin chain for synthesis. As shown in Figure \ref{fig:bg_inpainting}, our method successfully removes the target objects and inpaints the regions of the removed objects with reasonable backgrounds. Both appearance and motion of the inpainted video sequences are physically plausible.  

\section{Conclusion}

In this paper, we propose an energy-based spatial-temporal generative ConvNet model for  dynamic patterns, such as dynamic textures and action patterns. The model is in the form of deep spatial-temporal convolutional energy-based model where the energy function is defined by a bottom-up spatial-temporal ConvNet. The model corresponds to a spatial-temporal discriminative ConvNet classifier in the sense that the latter can be directly derived from the former. This property makes our model natural for modeling dynamic patterns. The learning of the model is achieved by an ``analysis by synthesis'' scheme: we sample the synthesized examples from the current model, usually by Markov chain Monte Carlo (MCMC), and then update the model parameters based on the difference between the observed training examples and
the synthesized examples. We show that the learning algorithm can be interpreted as an alternating mode seeking and mode shifting process, as well as an adversarial minimax optimization process. 

Our experiments show that the model can synthesize different types of realistic dynamic patterns, with well designed spatial-temporal ConvNet structures serving as energy functions.
Besides, it is possible to learn the model from videos with occluded pixels or missing frames. This can be achieved by adopting an extra Langevin dynamics starting from the corrupted training video sequences to recover the missing information. The resulting learning, sampling, and recovery algorithm is useful for unsupervised video inpainting, which includes video recovery (e.g., recovering missing pixels of frames of a video) and background inpainting (e.g., removing undesired moving object from a video). We show that learning from incomplete training data can provide an objective criterion to evaluate generative models of dynamic patterns.

\ifCLASSOPTIONcompsoc
  \section*{Acknowledgments}
\else
  \section*{Acknowledgment}
\fi

We gratefully acknowledge the support of NVIDIA Corporation with the donation of the Titan Xp GPU used for this research. The work is supported by ONR MURI N00014-16-1-2007, and  DARPA ARO W911NF-16-1-0579.

\ifCLASSOPTIONcaptionsoff
  \newpage
\fi

\bibliographystyle{IEEEtran}
\bibliography{mybibfile}

\end{document}